\pdfoutput=1
\documentclass[lettersize,journal,twoside]{IEEEtran}
\usepackage{amsmath,amsfonts}
\usepackage{algorithmic}
\usepackage{algorithm}
\usepackage{array}
\usepackage{textcomp}
\usepackage{stfloats}
\usepackage{url}
\usepackage{verbatim}
\usepackage{graphicx}
\usepackage{cite}
\usepackage{placeins}
\usepackage{stfloats}
\usepackage{caption}
\usepackage{subcaption}
\usepackage{pifont}
\usepackage{makecell}   % For multi-line cells
\usepackage{array}      % For fixed width columns
\usepackage[
    colorlinks=true,
    linkcolor=blue,
    citecolor=blue,
    urlcolor=blue
]{hyperref}
\usepackage{orcidlink}
\usepackage{setspace}

\newcommand{\quotes}[1]{``#1''}
\makeatletter
\renewcommand{\@cite}[2]{\textcolor{blue}{[{#1}\if@tempswa , #2\fi]}}
\makeatother

\newcommand{\eqbracketref}[1]{\textcolor{blue}{(}\ref{#1}\textcolor{blue}{)}}

\begin{document}

%\title{AutoCam: Autonomous Control of an Auxiliary Viewpoint for Surgical Robots}
%\title{AutoCam: Two-stage Path Planning for Auxiliary Camera Integration in Surgical Robots}
%\title{AutoCam: Development and Validation of a Two-stage Controller for an Auxiliary Camera in Surgical Robots}
%\title{AutoCam: Autonomous Camera Path Planning for Auxiliary Viewpoint in Surgical Robots}
%\title{AutoCam: Development and Validation of a Dual-Mode Path Planner for an Autonomous Auxiliary Camera in Surgical Robots}
\title{AutoCam: Hierarchical Path Planning for an Autonomous Auxiliary Camera in Surgical Robotics}

\author{Alexandre Banks\textsuperscript{\orcidlink{0009-0000-9617-4715}}, Randy Moore \textsuperscript{\orcidlink{0000-0003-3866-5246}}, Sayem Nazmuz Zaman \textsuperscript{\orcidlink{0009-0000-2554-2146}}, Alaa Eldin Abdelaal\textsuperscript{\orcidlink{0000-0002-0770-2940}}, and
Septimiu E. Salcudean\textsuperscript{\orcidlink{0000-0001-8826-8025}}, \textit{Life Fellow, IEEE}
        % <-this % stops a space
\thanks{This work was supported by the NSERC Canada Graduate Scholarships, the NSERC Discovery Grant, and the C.A. Laszlo Biomedical Engineering Chair held by Professor Salcudean. \textit{(Corresponding author: Alexandre Banks, abanksga@unb.ca)}}
\thanks{A. Banks and R. Moore contributed equally to this work.}
\thanks{A. Banks, R. Moore, S.N. Zaman and S.E. Salcudean are with the University of British Columbia, Vancouver, BC V6T 1Z4, Canada. A. E. Abdelaal is with Stanford University, Stanford, CA 94305, United States.}
\thanks{Supplementary downloadable material: \href{https://github.com/AlexandreBanks6/AutoCam_Supplemental_Files.git}{AutoCam\_Supplemental\_Files.git}.}
\thanks{Open-source implementation on the dVRK: \href{https://github.com/AlexandreBanks6/RCL_Autocam_Public.git}{RCL\_Autocam\_Public.git}}
}

% The paper headers
\markboth{A. B\MakeLowercase{anks} \MakeLowercase{and} R. M\MakeLowercase{oore} \MakeLowercase{\textit{et al.}}: AutoCam: Hierarchical Path Planning for an Autonomous Auxiliary Camera in Surgical Robotics}%
{A. B\MakeLowercase{anks} \MakeLowercase{and} R. M\MakeLowercase{oore} \MakeLowercase{\textit{et al.}}: AutoCam: Hierarchical Path Planning for an Autonomous Auxiliary Camera in Surgical Robotics}

% Remember, if you use this you must call \IEEEpubidadjcol in the second
% column for its text to clear the IEEEpubid mark.

\maketitle
\thispagestyle{empty} %Gets rid of header on first page
\begin{abstract}
Incorporating an autonomous auxiliary camera into robot-assisted minimally invasive surgery (RAMIS) enhances spatial awareness and eliminates manual viewpoint control. Existing path planning methods for auxiliary cameras track two-dimensional surgical features but do not simultaneously account for camera orientation, workspace constraints, and robot joint limits. This study presents AutoCam: an automatic auxiliary camera placement method to improve visualization in RAMIS. Implemented on the da Vinci Research Kit, the system uses a priority-based, workspace-constrained control algorithm that combines heuristic geometric placement with nonlinear optimization to ensure robust camera tracking. A user study (N=6) demonstrated that the system maintained 99.84\% visibility of a salient feature and achieved a pose error of 4.36 $\pm$ 2.11 degrees and 1.95 $\pm$ 5.66 mm. The controller was computationally efficient, with a loop time of 6.8 $\pm$ 12.8 ms. An additional pilot study (N=6), where novices completed a Fundamentals of Laparoscopic Surgery training task, suggests that users can teleoperate just as effectively from AutoCam's viewpoint as from the endoscope's while still benefiting from AutoCam's improved visual coverage of the scene. These results indicate that an auxiliary camera can be autonomously controlled using the da Vinci patient-side manipulators to track a salient feature, laying the groundwork for new multi-camera visualization methods in RAMIS.

\end{abstract}

\begin{IEEEkeywords}
robot-assisted surgery, da Vinci, stereoscopic imaging, autonomy, machine vision, autonomous robotic surgery
\end{IEEEkeywords}

\section{Introduction}
\IEEEPARstart{R}{obot} assisted minimally invasive surgery (RAMIS) has been adopted in over 60 countries \cite{rivero2023robotic} and is shown to reduce postoperative blood loss, shorten hospitalization times, and enable tremor filtering and enhanced dexterity \cite{hussain2014use,moorthy2004dexterity}. Most surgical robots, including the da Vinci (Intuitive Surgical, Inc.) and Hugo (Medtronic, Inc.) systems, have a single endoscopic camera (ECM) restricted to rotate about the remote center of motion (RCM) at the incision site \cite{avinash_pickup_2019}. Having only one viewpoint with limited maneuverability compromises global awareness of the surgical scene \cite{pandya2014review} and impedes surgical workflow when the endoscope is occluded \cite{eslamian_development_2020,velasquez2016preliminary,avinash_pickup_2019}. 

Incorporating an additional camera into RAMIS can provide different stereo perspectives across the viewing field, which improves perception, spatial awareness, and understanding of anatomical scale \cite{praveena2022understanding,urey2011state,su_multicamera_2021}. This enhanced contextual knowledge is especially important when teleoperating a robot in demanding environments \cite{chae2024divergent} such as in RAMIS. 

%\IEEEpubidadjcol %(This is start of second column, need this here to add the copyright)
Dual-view systems increase accuracy during surgically realistic training tasks \cite{abdelaal_multi-camera_2020} and having an additional perspective while watching training videos improves surgical performance by 29\% in trainees \cite{wang2019multiple}. Moreover, auxiliary viewpoint systems can lead to improved 3D reconstruction of surgical scenes \cite{tamadazte2015multi,su_deep_2022} which is particularly important for the integration of machine learning and augmented reality into RAMIS \cite{fluckiger_automatic_2025}. With medical error being the third leading cause of death in the US \cite{makary_medical_2016}, there is a need for auxiliary vision systems in surgery that can enhance situational awareness and enable continuous viewing of salient features in the surgical field.

To effectively use an additional viewpoint, the camera must be actively adjusted to view the surgical scene. While a surgeon could directly control the auxiliary camera, this unnecessarily adds to their workload of managing both surgical tools and the endoscope \cite{col_scan_2020, pandya2014review}. Studies show that completing additional tasks in surgery hinders performance on the primary task, particularly for novice surgeons \cite{zheng_measuring_2010}. This means that for an auxiliary camera to be integrated into RAMIS it is essential that the burden on the surgeon is not increased.

Another option is to delegate auxiliary camera control to a member of the surgical team, but this requires specialized training \cite{mariani_experimental_2020} and extra equipment in addition to the surgical robot. The availability of operating room (OR) personnel with robotic training is already limited \cite{randell2019factors} and robotic procedures are typically performed by a single teleoperator, making a dedicated assistant for auxiliary camera control impractical. Furthermore, manual camera guidance can cause surgical instruments to be out of the field of view (FoV) and longer procedure times \cite{elazzazi_natural_2022}. Challenges in manual control necessitate an autonomously driven auxiliary camera that reduces OR burden and allows surgeons to focus on instrument manipulation while receiving enhanced, multi-view, information. 

Much of the prior work has focused on hardware designs for auxiliary setups. One study used arrays of articulated miniature robots entering through natural orifices \cite{tortora2014array},
however, these do not integrate with the da Vinci Surgical System (dVSS) or autonomously track salient features. Other methods employing cable- or magnetically-driven cameras \cite{simi2011magnetic,rivas2017smart} are limited to movement along the abdominal wall, require a surgical assistant, and lack full 6 degree-of-freedom (6DOF) orientation and translation (pose) motion. Flexible imaging probes have been used in addition to the ECM \cite{velasquez2012development,velasquez2016preliminary} and multiple miniature cameras have been added to the end of an endoscope to expand the FoV \cite{tamadazte2015multi}. These methods, however, require manual telemanipulation and do not incorporate autonomy.

Although autonomous camera control can decrease distractions for the surgeon, minimize procedure time, and reduce surgical effort \cite{moccia_autonomous_2023,salcudean2022robot}, little research has combined autonomy with additional viewpoints. Most research on automatic auxiliary camera control has been limited to simulations and does not consider critical deployment factors such as orientation control and bounds on end-effector motion. A learning-from-demonstration approach was used to pan a camera during surgical debridement \cite{ji2018learning}, but motions were constrained to a 2D plane and did not consider workspace constraints. A series of articles \cite{subedi_smoothness_2023,su_deep_2022,su_multicamera_2020,su_multicamera_2021} developed a reinforcement learning (RL)-based approach to guide six camera viewpoints during RAMIS. Their approach is promising for scene reconstruction, multi viewpoint alignment, and tool-tip position tracking; however, the cameras were constrained to move along the plane of the abdominal wall without using all 6DOF. Additionally, independently controlling 6 abdominally-mounted cameras would require extra hardware not compatible with current RAMIS systems. These RL approaches were also only tested in simulation and did not consider robot joint limits. Another approach integrated kinematic constraints with autonomous control of a microscopic camera \cite{lin_autonomous_2024} and achieved 94.1\% accuracy in maintaining surgical instruments within the FoV. This deep-learning based approach, however, does not consider orientation of the salient feature, controlling only the camera's pan. In addition \cite{lin_autonomous_2024} loses depth information by only using a monocular camera, and the method was only validated on 2D trajectories. %To enable repeatable 6DOF camera motion, a pick-up camera was developed with a \quotes{stable grasp} coupling to the da Vinci ProGrasp surgical instrument \cite{avinash_pickup_2019}. 
%Since this initial work, our group has explored the effect of varying disparity of the pick-up stereo camera on depth perception \cite{avinash_evaluation_2020}, and have shown that a stationary auxiliary camera increases accuracy and completion speed of a dry-lab surgical training task by more than 25\% \cite{abdelaal_multi-camera_2020}. 
In Abelaal et al. \cite{abdelaal_orientation_2022}, autonomous 6DOF control of a pick-up camera %originally developed in \cite{avinash_pickup_2019} 
was proposed. However, the autonomous framework was only tested in simulation, tracking errors were not quantified, and the approach did not consider robot joint limits, robot calibration, or obstacle avoidance. Furthermore, it was not developed for the dVSS --- the most widespread surgical robot with over 6500 units worldwide \cite{rivero2023robotic}. 
To address these gaps, we present AutoCam: a framework for autonomous auxiliary camera control that considers joint and workspace constraints and enables full 6DOF positioning. AutoCam %integrates with the da Vinci system, and 
can be deployed on any da Vinci Research Kit (dVRK) using our open-source code: \href{https://github.com/AlexandreBanks6/RCL_Autocam_Public.git}{RCL\_Autocam\_Public.git}. 

Our goal is to %overcome these sim-to-real challenges and 
develop a markerless, real-time control approach for easy integration of a secondary viewpoint into RAMIS. To achieve this, our paper contributes the following: 1) a novel hierarchical auxiliary camera placement strategy that combines a naive geometric placement with a nonlinear optimization step to track salient surgical features; 2) workspace and joint limit constraints integrated into the control architecture; 3) an open-source implementation on the dVRK; 4) auxiliary camera stereo calibration and video processing; 5) a comprehensive error quantification of the system's 2D FoV and 6DOF tracking through a user study; and 6) a pilot study (N=6) evaluating the impact of AutoCam on novice performance. 
%This work aligns with the long-term vision of the RAMIS community to develop semi-autonomous surgical robots and automatic path planning algorithms for camera control \cite{millan_scoping_2021,d2021accelerating}.
Through this research, we aim to reduce occlusions and enhance visual-spatial perception during RAMIS while limiting unnecessary burden on the surgeon.

\section{Methods}
During an operation, the surgeon’s concentration is captured by various subtasks % from making incisions to closing each suture. These subtasks
which involve focusing on a salient feature (e.g. the incision site or tool location). We propose a markerless method in which one robotic manipulator autonomously positions an additional camera to track a salient feature, while the remaining manipulators are used to perform the surgical task. We focus on the single-handed wire-chaser task, an exercise from the Fundamentals of Laparoscopic Surgery (FLS) that is predictive of surgeon performance in the operating room \cite{aghazadeh2016performance, jarc2015face}, and represents elements of laparoscopic procedures \cite{peters2004development}. To provide an additional viewpoint that integrates with robotic surgical systems, we implement a framework building on the simulated work of Abdelaal et al. \cite{abdelaal_orientation_2022} where the ring of the FLS task is tracked as a salient feature. Our novel hierarchical controller aligns the camera's optical axis with the normal vector of the salient feature and keeps the feature at the center of view. Although we assume an orthogonal camera orientation is preferred, this viewing angle has proven useful in the wire-chaser task \cite{abdelaal_orientation_2022} and our method could be adapted to accommodate different viewing angles.

Section \ref{subsec:autocam_control_scheme} outlines path planning, robot calibration, and workspace constraints. Section \ref{subsec:teleoperation} covers hand-eye coordination when using the auxiliary camera and Section \ref{subsec:video_processing} details the video processing framework. ${}^{x}\boldsymbol{T}_{y}$ denotes a homogeneous mapping of 3D points $\boldsymbol{p}^y$ in coordinate system $\{\boldsymbol{o}_{y}, \underline{\boldsymbol{C}}_{y}\}$ (origin, frame) to points $\boldsymbol{p}^x$ in coordinate system $\{\boldsymbol{o}_{x}, \underline{\boldsymbol{C}}_{x}\}$. %AutoCam is available open-source at: \href{https://github.com/AlexandreBanks6/RCL_Autocam_Public.git}{RCL\_Autocam\_Public.git}.

\subsection{Autonomous Camera Control Scheme}
\label{subsec:autocam_control_scheme}
%general placement strategy 

\begin{figure}[h!]
\begin{center}    \includegraphics[width=\columnwidth,trim={0.6cm 0cm 0.15cm 0.25cm},clip]{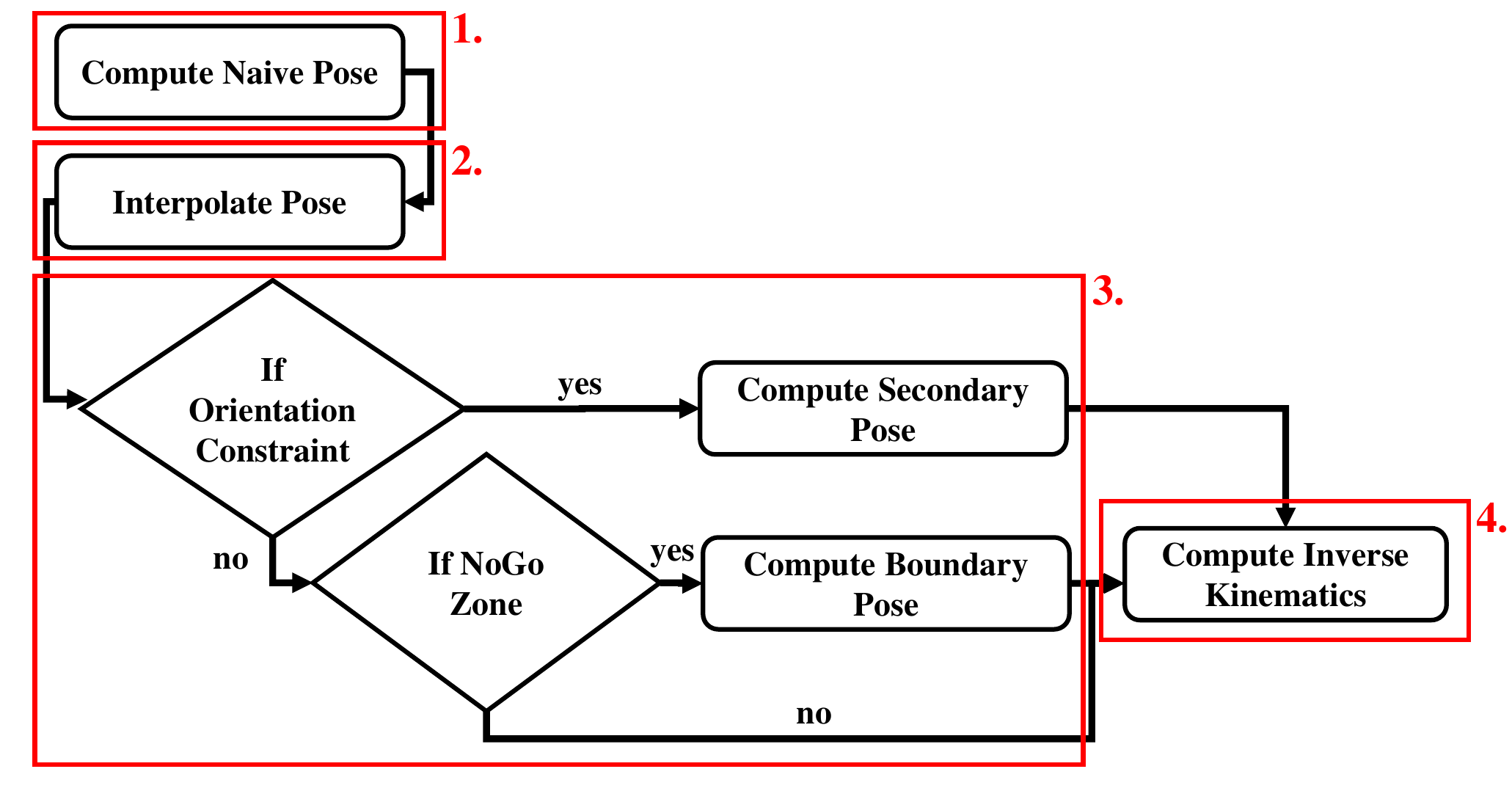}
    \caption{AutoCam path planning framework.}    \label{fig:Viewing Framework Controller Implementation}
\end{center}
\end{figure}

AutoCam's path planning framework illustrated in Fig.~\ref{fig:Viewing Framework Controller Implementation} consists of: 1) computation of a naive camera pose based on feature information and geometry (geometric-based placement), 2) interpolating the trajectory to the naive pose, 3) handling workspace constraints in Cartesian space, and 4) solving inverse kinematics with joint limit constraints. This is a \quotes{hierarchical} approach in that an unconstrained optimization is performed with the naive pose, which is then refined with a non-linear optimization if the constraints are violated.

\subsubsection{ Naive Geometric-Based Placement}
\label{subsubsec:gemoetric_based_placement}
\begin{figure}[h!]
\begin{center}    \includegraphics[width=\columnwidth,trim={0.5cm 9.8cm 2.8cm 0.4cm},clip]{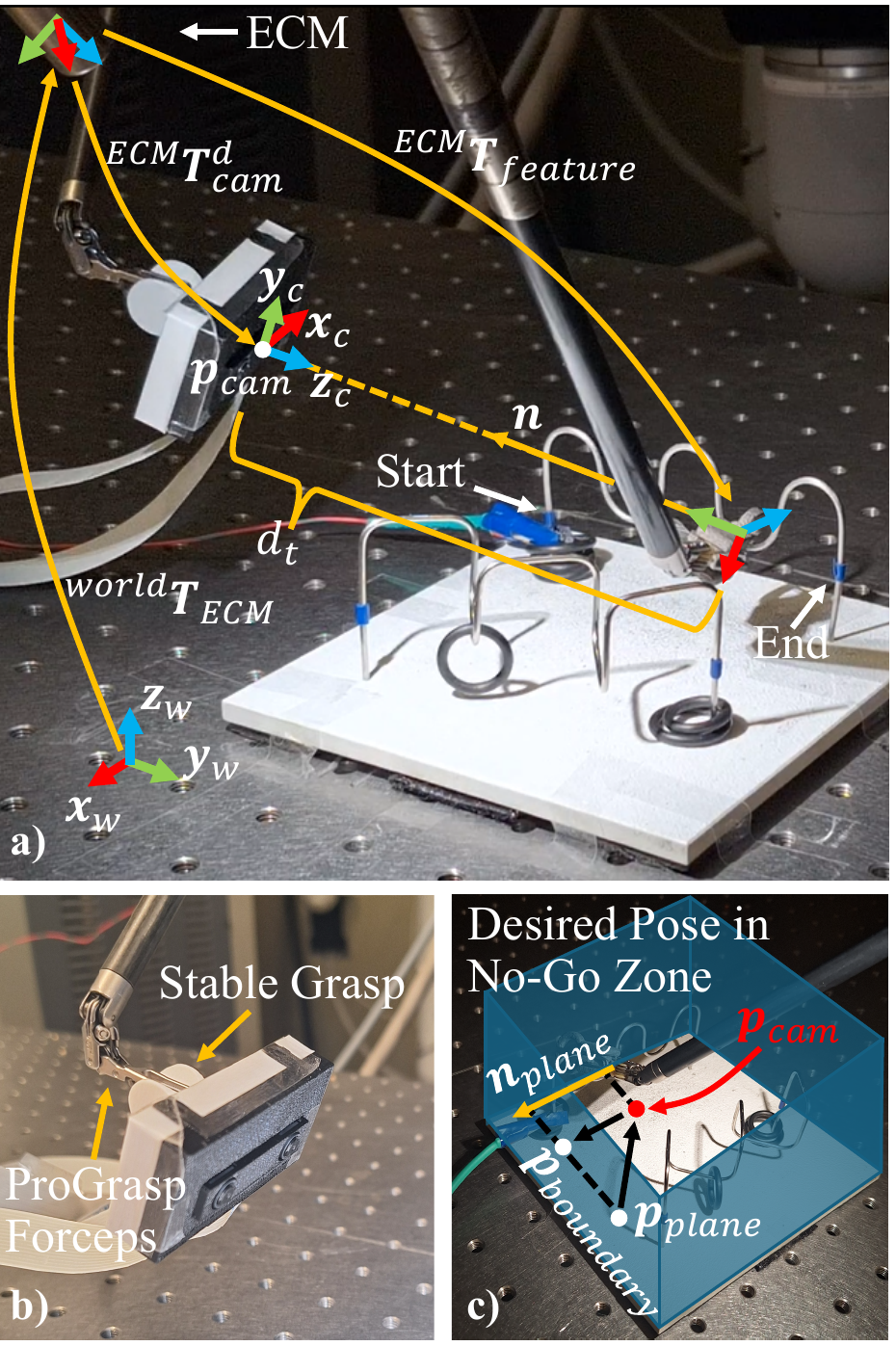}
    \caption{\textbf{a)} Homogeneous transformations and vectors for geometric-based camera placement. Transformations ${}^{ECM}\boldsymbol{T}_{cam}^d$ and ${}^{ECM}\boldsymbol{T}_{feature}$ map the endoscopic camera frame to the autonomous camera and salient feature frames, respectively. The vector $\boldsymbol{n}$ is normal to the tracked feature, and $d_{t}$ is the desired viewing distance between the feature and the camera. The start and end of the single-handed wire-chaser task are shown; \textbf{b)} The da Vinci\textsuperscript{\textregistered} ProGrasp Forceps holding the \quotes{Stable Grasp} mount to control the auxiliary camera; \textbf{c)} The No-Go Zone, illustrating a plane's normal vector $\boldsymbol{n}_{plane}$ and point $\boldsymbol{p}_{plane}$ on the plane. Also shown in the projection of the desired position $\boldsymbol{p}_{cam}$ onto the boundary as $\boldsymbol{p}_{boundary}$.}    \label{fig:GeometricController_AndMaterials}
\end{center}
\end{figure}

The first step in the hierarchical control framework is to compute a naive geometrically-based camera pose. To center the salient feature --- the ring of the surgical training task --- in the camera, the feature's pose, $^{ECM}\boldsymbol{T}_{feature}$, is tracked using the dVRK API \cite{kazanzides2014open} reported PSM pose and the rigid transform from the PSM to the feature centroid. We define its normal vector, $\boldsymbol{n}$, to lie along $^{ECM}\boldsymbol{T}_{feature}$'s $y$ axis as shown in Fig.~\ref{fig:GeometricController_AndMaterials}. The desired camera pose, $^{ECM}\boldsymbol{T}_{cam}^d$, is computed by first defining a viewing position, $\boldsymbol{p}_{cam}$, at a distance $d_t$ along the feature's normal vector, $\boldsymbol{n}$. Then, the camera's optical axis, $\boldsymbol{z}_{c}$, is aligned with the salient feature's normal vector. Finally, for a comfortable viewing experience, we constrain the $x$-axis of $^{ECM}\boldsymbol{T}_{cam}^d$ to lie within the $xy$ horizontal plane of the world coordinate system (the patient table). This is done by computing $\boldsymbol{x}_c$ as the cross product between the world $z$-axis, $\boldsymbol{z}_w$, and the camera optical axis, $\boldsymbol{z}_c$. This camera orientation is not uncommon in procedures; for example, the liver and pancreas should be horizontal in the surgical scene in upper abdominal procedures \cite{zheng2017your}. 
The camera's $y$-axis vector, $\boldsymbol{y}_c$ is the cross product of the $z$- and $x$-axes of the camera frame. 

%interpolation of poses 
Interpolation is performed on the desired pose, $^{ECM}\boldsymbol{T}_{cam}^d$, in Cartesian space before passing it to the joint controller. An intermediate setpoint is computed between the current and desired pose if the Euclidean distance between these two poses exceeds 1.5 cm. This setpoint is then used for joint space interpolation of position, velocity, and acceleration from the current camera configuration. The start conditions of this interpolation use the current velocity and acceleration, while the final conditions consist of zero velocity and acceleration. Defined joint trajectories are passed to the proportional integral derivative (PID) controllers for each joint. Interpolation was performed using the Collaborative Robotics Toolkit (CRTK) and the PID control using the saw/cisst stack \cite{9287933}. Path planning was designed without explicit dynamic modelling to maintain computational efficiency and leverage the robot's existing PID control architecture. Future designs could impose dynamic constraints on the computed paths; however, our method was found to work for the relatively small motions and low joint speeds of the surgical instruments.%This can be explained by the relatively slow joint speeds required of the autonomous camera arm making higher-order terms in robotic manipulator dynamic models, such as the Euler-Lagrange, negligible.  

\subsubsection{ Arm-calibration } 
The da Vinci patient-side manipulators are 13 DOF robotic arms, with several joints being cable-driven. Such long kinematic chains, combined with deformation due to external forces, can result in tooltip localization error when using forward kinematics alone \cite{kwartowitz2006toward}. As a result, we present a markerless calibration method to localize the autonomous camera arm with the arm being teleoperated for accurate kinematic-based tracking. To accomplish this, paired points of the arms are collected by touching their tooltips together. The relationship between the two arms is modeled as a homogeneous transform which is solved by minimizing the mean absolute error between the reprojected points. This kind of markerless calibration can be performed intraoperatively. 

\subsubsection{ Workspace Constraints}
Workspace constraints are imposed to facilitate robust tracking of the salient feature and to generalize to an intraoperative scenario by preventing collisions with tissue. These constraints consist of virtual fixtures defined by the teleoperating surgeon, who outlines a No-Go Zone by moving the tip of the patient-side manipulator. In our study, the No-Go Zone is defined as a rectangular prism enclosing the wire-chaser task, but this definition can be generalized to any shape. 
The surgeon defines the base and height as a set of five points using the patient-side manipulator. A cube is then fit to these points, with each face, $i$, of the cube defined by a normal vector, $\boldsymbol{n}_{plane, i}$ facing outwards, and a point $\boldsymbol{p}_{plane, i}$ lying on the plane. The desired pose, $^{ECM}\boldsymbol{T}_{cam}^d$, with position $\boldsymbol{p}_{cam}$ was determined to be in the No-Go Zone if any of the projections of vectors defined by $(\boldsymbol{p}_{cam}-\boldsymbol{p}_i)$ onto the normal vector $\boldsymbol{n}_{plane,i}$ were negative.

If a desired pose is in the No-Go Zone, a \emph{Boundary-Pose} is computed that lies on the No-Go Zone. This pose is computed by first finding the plane of the No-Go Zone closest to $^{ECM}\boldsymbol{t}_{cam}^d$ and extracting the plane's normal vector $\boldsymbol{n}_{plane}$ and a point $\boldsymbol{p}_{plane}$ in the plane. The shortest distance from $^{ECM}\boldsymbol{t}_{cam}^d$ to the plane is computed by projecting the vector from $\boldsymbol{p}_{plane}$ to $^{ECM}\boldsymbol{t}_{cam}^d$ onto the normal vector and translating $\boldsymbol{p}_{cam}$ along this direction as shown in equation \eqbracketref{Eq: p_boundary}: 
\begin{equation}\label{Eq: p_boundary}
    \boldsymbol{p}_{boundary} = [(\boldsymbol{p}_{cam}- \boldsymbol{p}_{plane}) \cdot \boldsymbol{n}_{plane}] \, \boldsymbol{n}_{plane}
\end{equation}
The orientation of the camera, $^{ECM}\boldsymbol{R}_{cam}^d$, is then recomputed at this location to point at the feature's centroid. 
The orientation constraint (Fig. \ref{fig:Viewing Framework Controller Implementation} box 3) is imposed to keep the camera on one side of the task reducing collisions with tools and tissue. A preferred side is defined, and violation of this constraint is monitored by the angle between the cameras viewing axis and the $x$-axis of the world. 

\subsubsection{Naive Inverse Kinematics}
\label{subsubsec:joint_limits}

The unconstrained inverse kinematics (IK) native to the dVRK is used to compute naive joint angles, $\boldsymbol{q}_{d}$, required to position the camera to track the salient feature. This IK is performed 
using Newton's method, which is computationally efficient when a solution exists within a well-conditioned region. It relies on quadratic convergence and is highly effective when starting near a solution. However, it can fail if it is far from a feasible solution or when singularities are encountered. In situations where IK fails, or $\boldsymbol{q}_d$ violates the lower and upper robot joint limits ($\boldsymbol{q}_l$ and $\boldsymbol{q}_u$), a more robust constrained nonlinear optimization is performed. Unlike the naive approach, this optimization-based IK explicitly incorporates joint limit constraints and other controller constraints as weighted costs. Our approach aligns with the general intuition that robotic controllers should use fast solvers first and reserve more complex solvers for exceptional cases. 

\subsubsection{Constrained Inverse Kinematics Solver}
\label{subsec:constrained_IK}
Should the desired joint vector, $\boldsymbol{q}_{d}$, computed by Newton's method violate joint limits, a new joint vector, $\boldsymbol{q}_{o}$, is found using constrained optimization. This optimization has three cost terms: 1) a position cost, $C_{ps}$, enforcing the camera to the desired position, 2) an orientation cost, $C_{or}$, enforcing the desired view angle of the camera, and 3) a distance cost, $C_{dis}$, enforcing the distance between the camera and the ring to a desired threshold, $ d_{t}$. These cost terms are given in the objective function \eqbracketref{optim_eq1}.
\begin{equation}
\label{optim_eq1}
\begin{aligned}
f_{loss}(\boldsymbol{q}_{o}) = 
&w_1h\big\{C_{ps}(\boldsymbol{q}_{o}, {}^{ECM}\boldsymbol{t}_{cam}^{d}),\delta_1\big\} + \\
&w_2h\big\{C_{or}(\boldsymbol{q}_{o}, \boldsymbol{c}, \boldsymbol{y}_{w}),\delta_2\big\} + \\
&w_3h\big\{C_{d}(\boldsymbol{q}_{o}, {}^{ECM}\boldsymbol{T}_{feature}, d_{t}),\delta_3\big\}
\end{aligned}
\end{equation}

$C_{ps}$ penalizes the Euclidean norm between the camera position returned by the optimization, ${}^{ECM}\boldsymbol{t}_{cam}^{o}$, and the desired camera position computed geometrically, ${}^{ECM}\boldsymbol{t}_{cam}^{d}$,i.e., $C_{ps}=||{}^{ECM}\boldsymbol{t}_{cam}^{o}-{}^{ECM}\boldsymbol{t}_{cam}^{d}||$. The translation ${}^{ECM}\boldsymbol{t}_{cam}^{o}$ is computed in \eqbracketref{optim_eq2} as a function of $\boldsymbol{q}_{o}$ using the forward kinematics, $f_{k}$, of the PSM, where  %using the Denavit Hartenberg parameters described in \cite{hwang2020superhuman}.
${}^{PSM}\boldsymbol{T}_{cam}$ is a rigid transform from the end-effector to the camera frame.
\begin{equation}
\label{optim_eq2}
\begin{aligned}
    {}^{ECM}\boldsymbol{T}_{cam}^{o}&=\{{}^{ECM}\boldsymbol{R}_{cam}^{o},{}^{ECM}\boldsymbol{t}_{cam}^{o}\}\\
    &=f_{k}(\boldsymbol{q_{o}}) \, \cdot \,{}^{PSM}\boldsymbol{T}_{cam}
\end{aligned}
\end{equation}

%The viewing vector angle error is computed by first finding a vector, $\boldsymbol{c}$, subtending from the camera's coordinate frame origin to the feature's coordinate frame origin. The angular difference between $c$ and the camera's optical axis, $\boldsymbol{z}_{c}$, was taken as the viewing angle error.

To ensure proper orientation of the autonomous view, $C_{or}$ in \eqbracketref{optim_eq1} enforces two conditions: 1) the camera points toward the salient feature and 2) the camera's $x$-axis, $\boldsymbol{x}_c$, lies within the $xy$ horizontal plane of the world coordinate system. $C_{or}$ is computed through a compound cost \eqbracketref{optim_eq3:A}. The first term, $C_v$, penalizes misalignment between the auxiliary camera's optical axis, $\boldsymbol{z}_{c}^{o}$, and a vector, $\boldsymbol{c}$, subtending from the auxiliary camera's origin to the feature's origin. $C_v$ uses a 1-cosine similarity measure \eqbracketref{optim_eq3:D}. The second term, $C_{per}$, penalizes deviation from an angle of $\pi/2$ between the auxiliary camera's x-axis, $\boldsymbol{x}_{c}^{o}$, and the world's vertical axis, $\boldsymbol{z}_w$. $C_v$ is pre-multiplied by $1/2$ to ensure the range is the same as $C_{per}$: $[0,1]$. $w_4$ and $w_5$ are regularization terms of $C_{or}$, with values given in Appendix~\ref{appendix:parametervals}. $\boldsymbol{x}_{c}^{o}$ and $\boldsymbol{z}_{c}^{o}$ are the $\boldsymbol{x}$ and $\boldsymbol{z}$ positional components of ${}^{ECM}\boldsymbol{T}_{cam}^{o}$ computed in \eqbracketref{optim_eq2}.
\begin{subequations}\label{optim_eq3}
\begin{align}
C_{or}(\boldsymbol{q}_{o}, \boldsymbol{c},\boldsymbol{y}_{w})&=w_4\frac{1}{2}C_{v}(\boldsymbol{z}_{c}^{o},\boldsymbol{c})+w_5C_{per}(\boldsymbol{x}_{c}^{o},\boldsymbol{z}_{w}) \label{optim_eq3:A}\\
C_{v}(\boldsymbol{z}_{c}^{o},\boldsymbol{c})&=1-f_{csim}(\boldsymbol{z}_{c}^{o},\boldsymbol{c})\label{optim_eq3:B}\\
C_{per}(\boldsymbol{x}_{c}^{o},\boldsymbol{z}_{w})&=|f_{csim}(\boldsymbol{x}_{c}^{o},\boldsymbol{z}_{w})|\label{optim_eq3:C}\\
f_{csim}(\boldsymbol{v}_1,\boldsymbol{v}_2)&=\frac{\boldsymbol{v}_1\cdot\boldsymbol{v}_2}{||\boldsymbol{v}_1|| \,||\boldsymbol{v}_2||}\label{optim_eq3:D}
\end{align}
\end{subequations}

The final cost term, $C_{d}$, in \eqbracketref{optim_eq1} enforces a desired distance, $d_{t}$, between the auxiliary camera and the feature being tracked, ${}^{ECM}\boldsymbol{T}_{feature}=\{{}^{ECM}\boldsymbol{R}_{feature},{}^{ECM}\boldsymbol{t}_{feature}\}$, such that:

\begin{equation}\label{eq: distance cost}
    C_d=||{}^{ECM}\boldsymbol{t}_{feature}-{}^{ECM}\boldsymbol{t}_{cam}^{o}||-d_t
\end{equation}

Huber loss, $h\{\cdot,\cdot\}$, is applied to each cost term in the objective function \eqbracketref{optim_eq1}. This robust evaluation function proposed by \cite{huber_robust_1964} is less sensitive to outliers in the cost evaluations. See Appendix~\ref{appendix:eqdefs} for the definition of $h\{\cdot,\cdot\}$.

Optimal joint parameters, $\boldsymbol{q}_{o}^*$, are solved using an optimization \eqbracketref{optim_eq4:A} subject to joint limit constraints given in \eqbracketref{optim_eq4:B}. 
\begin{subequations}\label{optim_eq4}
\begin{align}
\boldsymbol{q}_o^*=\underset{\boldsymbol{q}_o}{\text{arg min}}\{f_{loss}(\boldsymbol{q}_o)\}\label{optim_eq4:A}\\
\text{subject to } \boldsymbol{q}_l \leq\boldsymbol{q}_o \leq \boldsymbol{q}_u\label{optim_eq4:B}    
\end{align}    
\end{subequations}

Constraints $\boldsymbol{q}_l$ and $\boldsymbol{q}_u$ are the robot joint limits.
$f_{loss}(\boldsymbol{q}_{o})$ is minimized in real-time using a sequential least-squares quadratic programming algorithm \cite{kraft_algorithm_1994} deployed using the Drake \cite{tedrake_drake_2019} Python~3 binding of the open-source nonlinear optimization library NLopt~\cite{johnson_nlopt_nodate}. In our implementation, the optimizer runs for a maximum of 50 function evaluations and terminates when the objective function value changes less than a threshold of $5^{-4}$. The optimization is seeded with the robot's current joint parameters, $\boldsymbol{q}_{cur}$, to increase the likelihood that the solution to \eqbracketref{optim_eq4:A} is within a local neighborhood.

\subsection{Visual-Motor Aligned Teleoperation} 
\label{subsec:teleoperation}
%visual motor alignment
%teleoperation frame filtering and updating scheme 
%visual motor alignment may be left out given that we did not use it in our most recent user study 
Visual-motor misalignment occurs when the viewpoint is not aligned with the control frame from which teleoperation is performed. This is common in laparoscopic procedures where the surgeon operates tools from an endoscope not necessarily aligned with the tools themselves and is known to negatively impact operator performance \cite{chintamani2009improved}. To address visual-motor misalignment, we developed an approach where the da Vinci arms are teleoperated from the moving viewpoint of the auxiliary camera. This is implemented as position-control, where the orientation of the teleoperated arm with respect to the auxiliary camera frame matches the orientation of the master tool manipulator (MTM) with respect to the console display. The position of the auxiliary camera is updated when the MTM translates and is frozen by clutch presses.

%An additional setting of the autonomous camera was developed to minimize the effect of visual-motor misalignment while using the auxiliary camera for teleoperation. This approach allows the teleoperation of a da Vinci surgical robotic arm from the viewpoint of the auxiliary camera - that is, the control frame of the teleoperation is from the pose of the auxiliary camera, providing a more intuitive method for utilizing the auxiliary viewpoint.  
\subsection{Autonomous Camera Video Processing}
\label{subsec:video_processing}

\begin{figure}[h!]
\begin{center}    \includegraphics[width=0.95\columnwidth,trim={1cm 1.7cm 42cm 6.8cm},clip]{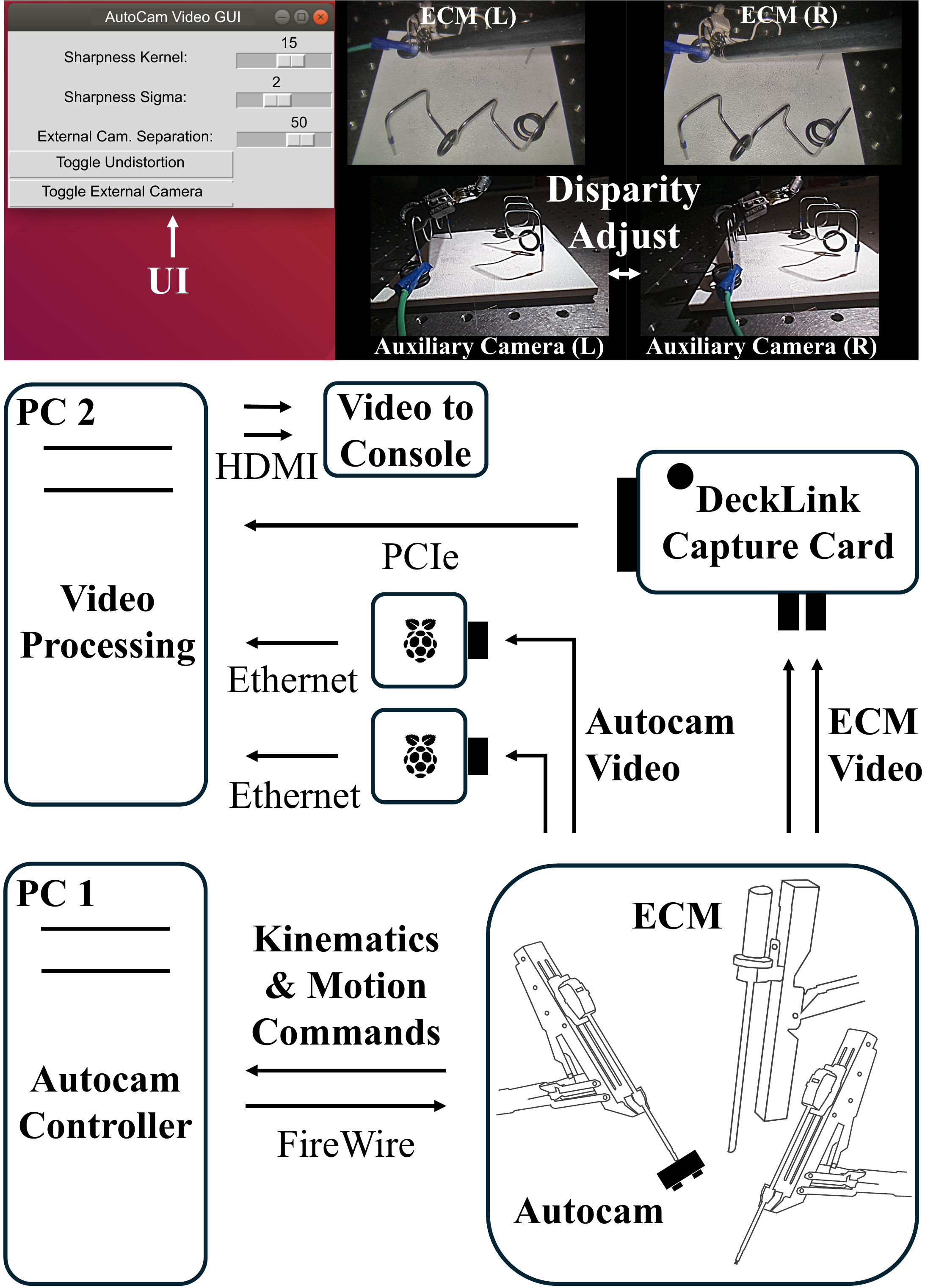}
    \caption{\textbf{Top:} The user interface (UI) and the console view with vertically concatenated ECM and autonomous camera frames. The UI allows adjustment of the on-screen separation (disparity) of the left/right autonomous camera frames. \textbf{Bottom:} Network diagram of all data streams. PC1 receives end-effector poses via the dVRK API, and sends commands over a FireWire connection to move the PSMs. The autonomous camera is interfaced by two Raspberry Pi\textsuperscript{\textregistered} controllers and video is sent to PC2 over ethernet for processing. The ECM video is captured and relayed to PC2 via a DeckLink capture card. Concatenated ECM and autonomous camera frames are sent to the left/right console viewports via an HDMI cable.}
    \label{fig:network_andGUI}
\end{center}
\end{figure}

The auxiliary camera consists of a pair of V2 Raspberry Pi\textsuperscript{\textregistered} Camera Modules and is integrated with the dVRK as shown in Fig.~\ref{fig:network_andGUI}b. Each camera connects to a Raspberry Pi 4 Model~B processor via a MIPI\textsuperscript{\textregistered} Display Serial Interface. Using the Robotic Operating System, each Pi sends RGB frames in real-time to a secondary PC over ethernet. Frames are received with a resolution of $1280\times960$ at 30 Hz. The secondary PC further processes AutoCam video frames using the Open Source Computer Vision Library (OpenCV) 4.8 in Python 3.

Frames captured by the autonomous stereo camera are corrected for radial and tangential lens distortion to improve image matching between the left/right views. Undistortion uses a pinhole camera model with intrinsics derived from the calibration method in \cite{zhang2000flexible}. For calibration, 41 images of a $9\times10$ checkerboard pattern were captured at varying poses. A two-stage camera algorithm is used in which: 1) monocular calibration computes each camera's intrinsics, and 2) stereo calibration minimizes projection error across both cameras and aligns the stereo pair into a unified 3D coordinate system.

Following undistortion, autonomous camera frames are sharpened using an image filtering technique from \cite{he2012guided}. The sharpened image, $\boldsymbol{S}$, is computed by finding the difference between the original image, $\boldsymbol{I}$, and its Gaussian blur, $\boldsymbol{G}$, such that $\boldsymbol{S}=(\alpha+1)\boldsymbol{I}-\alpha\boldsymbol{G}$. Our implementation set $\alpha$ to $1$.

As a final processing step, on-screen separation between the left/right autonomous camera frames is set depending on user preference. Camera disparity impacts depth perception \cite{avinash_evaluation_2020} and visual fusion \cite{pepper1984research}; therefore, we synthetically modulate disparity by adjusting the image separation in the console view with the interface in Fig.~\ref{fig:network_andGUI}a. Users adjust this separation by focusing on the surgical instrument's tool-tip and alternately opening and closing each eye. Disparity is adjusted until no perceptible shift occurs when switching between eyes.

Processed autonomous camera frames are vertically concatenated with the ECM video frames (Fig.~\ref{fig:network_andGUI}a.) and streamed to the dVRK \cite{kazanzides2014open}. The da Vinci Si endoscope frames are captured by the secondary PC using a Blackmagic Design\textsuperscript{\textregistered} DeckLink Duo 2 video capture card. The synchronized autonomous camera and ECM frames are displayed on left/right monitors in the dVRK console, providing an additional viewpoint of the surgical scene with 3D stereo vision.

\subsection{Experimental Design}
\label{subsec:experimental_design}

AutoCam was deployed on the dVRK and its ability to track the salient feature (ring) was quantified while human participants controlled the robot. We analyzed our autonomous camera approach by conducting three experiments: 1) characterizing overall system performance based on the reported pose of the dVRK API, 2) quantifying how close the feature of interest was to the center of the image using a visual tracking baseline, and 3) isolating the performance of the constrained IK solver. During each experiment, two different participants completed the single-handed wire-chaser task adapted from the FLS, giving a total of 6 trials. Each participant completed three sets of the task, which required moving a ring along a wire and back. Users were instructed to balance completion time with the number of collisions made with the wire. 

A separate pilot study with 6 participants was also conducted to understand AutoCam's impact on novice performance with the single-handed wire-chaser task. A within-subject design tested three conditions: 1) teleoperating from the stationary endoscope with only the endoscope view shown in the console (ENDO), 2) teleoperating from a stationary auxiliary camera (STAT) with both endoscopic and auxiliary views shown, and 3) teleoperating from the autonomous auxiliary viewpoint (AUT) with both views shown. Condition order was pseudo-randomized using a Latin Square approach \cite{avinash_evaluation_2020}, ensuring each permutation of conditions occurred equally across the population. For each condition, participants completed: 1) five practice sessions, 2) a two-minute break, 3) five evaluation trials where performance was recorded, and 4) another two-minute break to complete the NASA Task Load Index (NASA-TLX) Questionnaire \cite{hart_nasa-task_2006}. A piece of tape was fixed to the wire-chaser task (Fig. \ref{fig:GeometricController_AndMaterials}c.), and recording started/ended when the ring passed this marker. Collisions were detected via electrical contact between a nickel-coated ring and the task's wire as in \cite{banks2025setup}. An Arduino Mega sent collision data via serial to a PC, synchronizing it with instrument pose. 

Ethics approval was received from the University of British Columbia Behavioural Research Ethics Board under study number \#H24-02917. All participants included in the study provided informed consent using the approved consent forms.

\section{Results}

System errors are quantified in Section \ref{subsec:overall_system_performance}, and 2D centroid tracking is evaluated against a visual baseline in Section \ref{subsec:visual_centroid_tracking}. Section \ref{subsec:IK_performance} isolates constrained IK performance, and Section \ref{subsec:usability_analysis} presents a user study demonstrating AutoCam's impact on novice performance during the wire-chaser task.

\subsection{Overall System Performance}
\label{subsec:overall_system_performance}

Overall tracking performance was quantified on 952.4 seconds (N=75463 samples) of data from two participants with errors summarized in Table~\ref{table:overall_system_errors}. Viewing vector angle (\textit{VVA}) error was the angular difference between $\boldsymbol{c}$, the vector from the auxiliary camera's origin to the feature's origin, and $\boldsymbol{z}_{c}$, the auxiliary camera's optical axis. Feature distance (\textit{FD}) error was the difference between the desired feature distance, $d_{t}=0.11$m, and the actual distance during tracking. Angular error in keeping the auxiliary camera parallel to the floor (\textit{PF}) was measured. Additionally, the angular (\textit{AN}) and positional (\textit{PN}) differences between the naive, geometrically computed pose and the actual commanded pose are given. Loop time (\textit{LT}) is the average time per controller iteration. Column 1 of Table~\ref{table:overall_system_errors} gives total errors, column 2 includes errors only when constraints were not encountered (\textit{WoC}), and column 3 shows errors while constrained-optimization IK is used (\textit{WC}).

\begin{table}[h!]
    \centering
    \caption{Overall System Errors}
    \label{table:overall_system_errors}
    \setlength{\tabcolsep}{5pt} % Reduce column padding
    \begin{tabular}{llll}

    \hline\hline
          & \makecell[l]{\textbf{All}} & \makecell[l]{\textbf{WoC}$^{[1]}$} & \makecell[l]{\textbf{WC}$^{[2]}$}\\
    \hline
    Samples & 75463 & 45543 & 29920\\
    $VVA$ (\textdegree) & 1.71 $\pm$ 1.54 & 1.60 $\pm$ 1.08 & 1.88 $\pm$ 2.03 \\
    $FD$ (mm) & 4.98 $\pm$ 5.00 & 3.01 $\pm$ 3.61 & 7.98 $\pm$ 5.31\\
    $PF$ (\textdegree)& 0.34 $\pm$ 0.21 & 0.37 $\pm$ 0.22 & 0.29 $\pm$ 0.17\\
    $AN$ (\textdegree)& 30.39 $\pm$ 37.68 & 6.58 $\pm$ 6.72 & 66.63 $\pm$ 36.54\\
    $PN$ (mm)& 16.43 $\pm$ 18.66 & 4.63 $\pm$ 3.03 & 34.39 $\pm$ 18.16\\
    $LT$ (ms)& 6.75 $\pm$ 12.80 & 1.33 $\pm$ 0.71 & 14.98 $\pm$ 17.33\\
    \hline
    \end{tabular}
    \vspace{0.05em} % Space between table and note
    \parbox{0.99\linewidth}{\footnotesize%
    \vspace{0.4em}
    %$^{[1]}$Errors over all samples.  
    $^{[1]}$Errors without workspace constraints (WoC) included.

    $^{[2]}$Errors only when workspace constraints (WC) are encountered.}
\end{table}

\begin{figure}[h!]
\begin{center}    \includegraphics[width=\columnwidth,trim={1.8cm 4.1cm 3.1cm 0.95cm},clip]{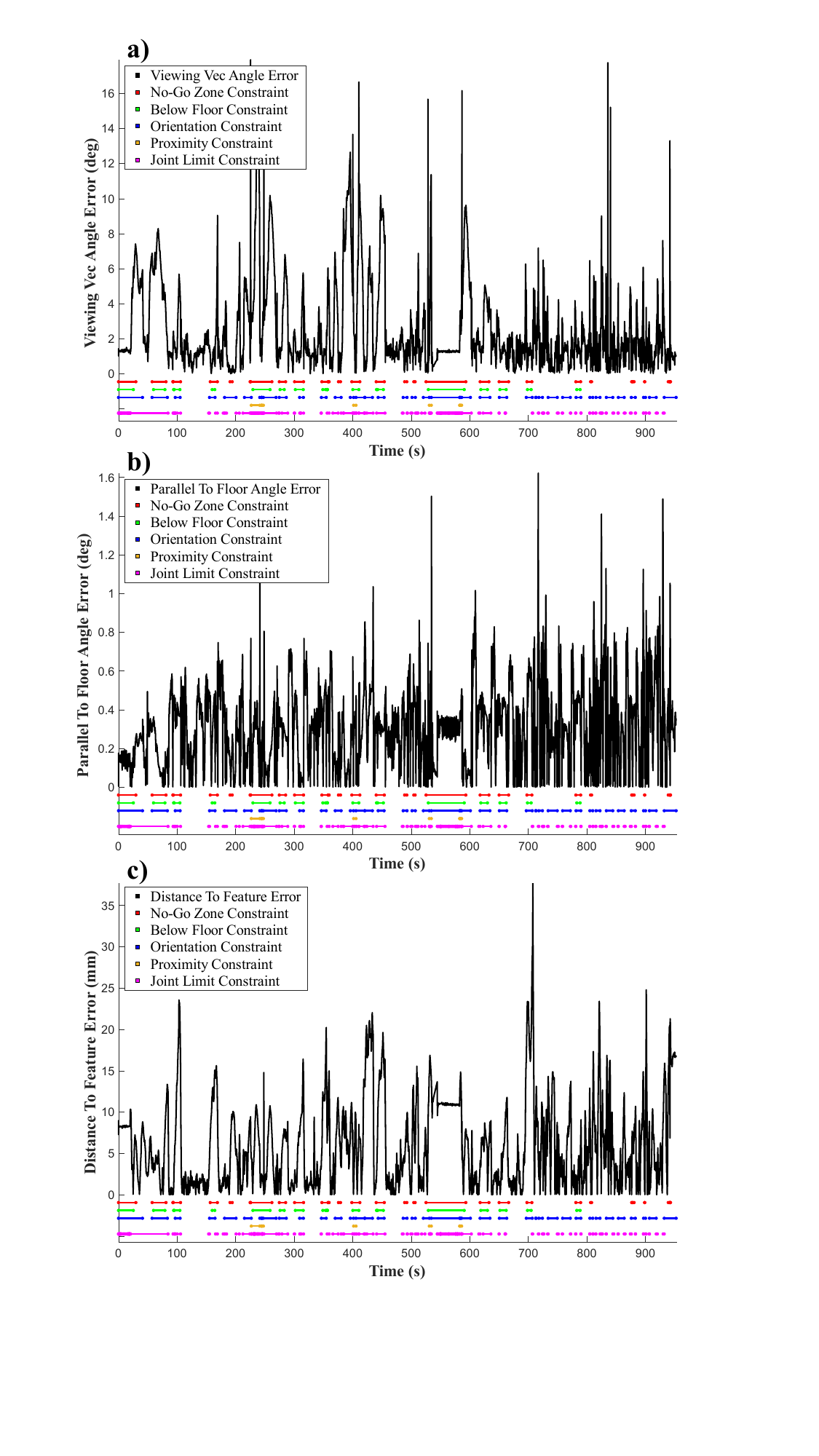}
    \caption{Absolute tracking errors (black) with bars (red, green, blue, organge, cyan) indicating when workspace and robot constraints were encountered. \textbf{a)} Angular distance between the auxiliary camera's optical axis and the vector connecting the its origin to the feature origin ($VVA$); \textbf{b)} Error between desired and actual camera-to-feature distance ($FD$); \textbf{c)} Angular error between camera and world horizontal axes ($PF$).}
    \label{fig:viewing_errors}
\end{center}
\end{figure}

Fig.~\ref{fig:viewing_errors} illustrates tracking errors under different workspace constraints. The \textit{No-Go Zone} and \textit{Below Floor} constraints occurred when the naive %geometrically computed 
pose moved outside the workspace. The \textit{Proximity Constraint} was encountered when the auxiliary camera was too close ($<$8cm) to the feature, and the \textit{Joint Limit Constraint} when the robot met a mechanical limit. 
%The \textit{Orientation Constraint} was triggered when the camera points towards the side of the scene that should not be traversed.

\begin{figure}[h!]
\begin{center}    \includegraphics[width=\columnwidth,trim={1.6cm 6.1cm 3cm 0.5cm},clip]{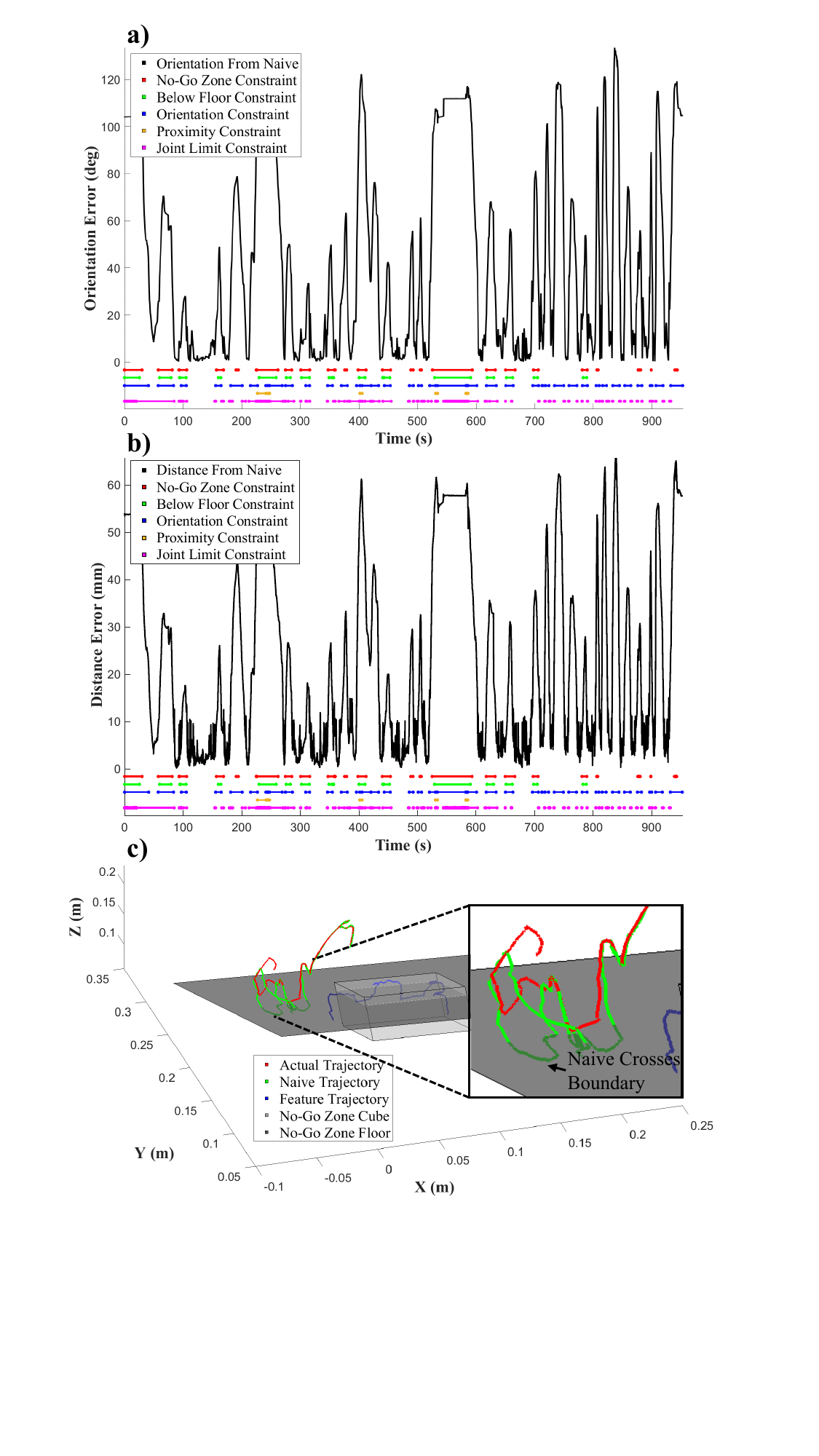}
    \caption{\textbf{a)} and \textbf{b)} Orientation and position error ($AN$ and $PN$) between the naive pose computed by the geometric-based placement and the actual pose from the hierarchical controller. Horizontal bars indicate constraint conditions; \textbf{c)} Example of the naive (green), actual (red), and feature trajectories (blue) with workspace constraints (grey) shown.}
    \label{fig:actual_to_naive}
\end{center}
\end{figure}

The difference between the naive pose and the actual commanded pose are illustrated in Fig.~\ref{fig:actual_to_naive}. Angular error (\textit{AN}) is shown in Fig.~\ref{fig:actual_to_naive}a., and position error (\textit{PN}) in Fig.~\ref{fig:actual_to_naive}b. Sharp peaks in the error signals arise when constraints are encountered, causing average errors to be higher ($AN$=30.39 $\pm$ 37.68\textdegree; $PN$=16.43 $\pm$ 18.66 mm). However, when samples with constraint conditions are removed, errors were lower ($AN$=6.58 $\pm$ 6.72\textdegree \;and $PN$=4.63 $\pm$ 3.03 mm). Fig.~\ref{fig:actual_to_naive}c. gives an example of the actual, naive, and feature trajectories with the workspace shown as a semi-transparent gray boundary. While the naive trajectory (green) crossed the workspace boundary, the commanded trajectory (red) did not. 

\subsection{Visual Centroid Tracking Experiment}
\label{subsec:visual_centroid_tracking}

\begin{figure*}[!t]
\begin{center}
    \includegraphics[width=\textwidth,trim={0.5cm 6.1cm 1cm 8.6cm},clip]{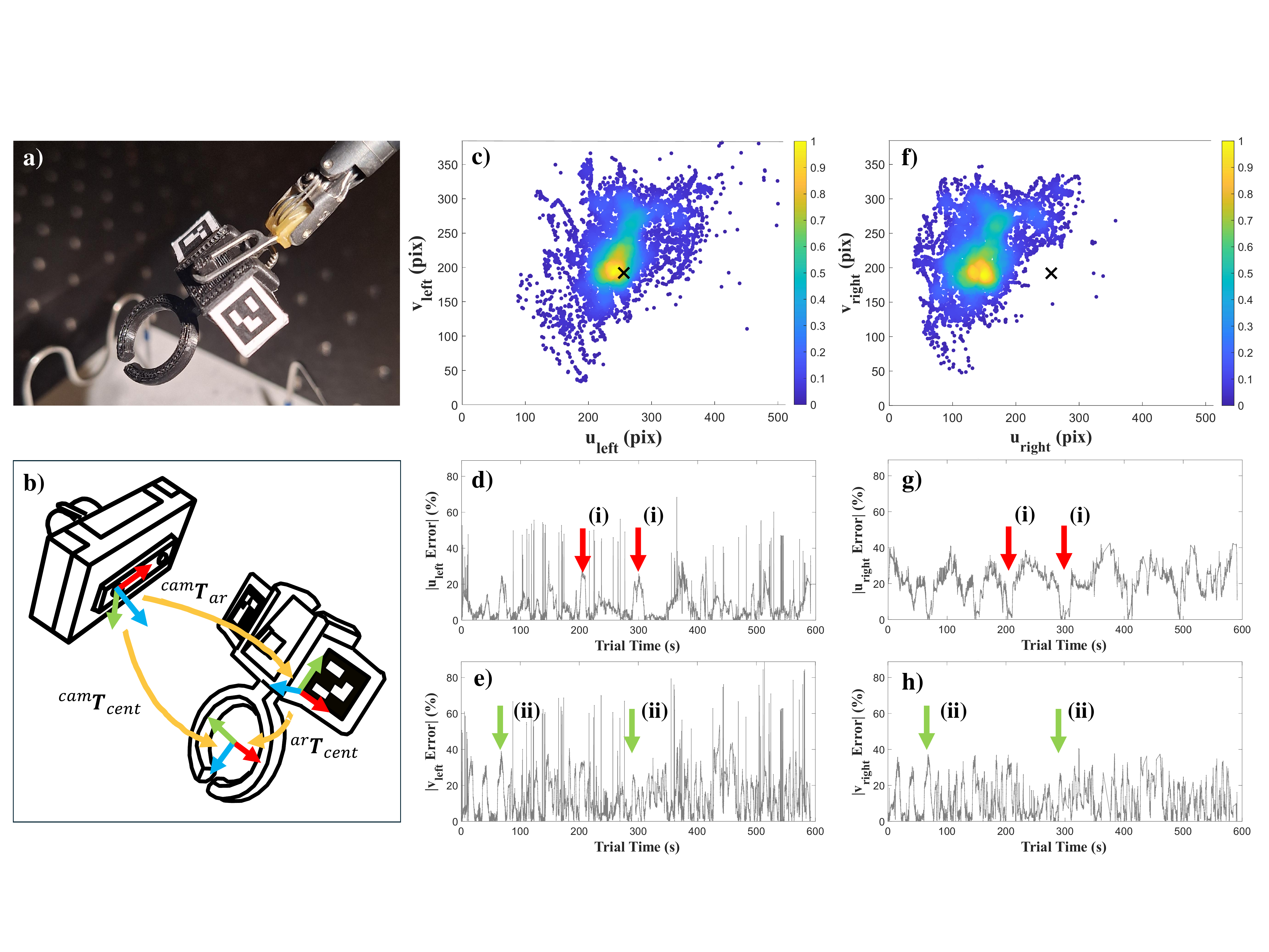}
    \caption{Results of tracking the feature's centroid. \textbf{a)} The ring feature with the modified \quotes{stable grasp} object; \textbf{b)} Coordinate transformations to compute the pose of centroid with respect to the auxiliary camera frame, ${}^{cam}\boldsymbol{T}_{cent}={}^{cam}\boldsymbol{T}_{ar}{}^{ar}\boldsymbol{T}_{cent}$; \textbf{c)} and \textbf{d)} are scatter density plots of the 2D centroid coordinates with \quotes{$\times$} being the center of the camera frame; \textbf{d)} and \textbf{e)} show the absolute value of the $(u,v)$ component-wise tracking error plotted against time for the left auxiliary camera; \textbf{g)} and \textbf{h)} show the $(u,v)$ component-wise tracking error for the right camera. The labels (i) (red arrows) and (ii) (green arrows) indicate error peaks at corresponding time instances.}
    \label{fig:centroid_tracking}
\end{center}
\end{figure*}

To ensure the feature is not drifting out of view and remains centered on the screen for the surgeon, we quantified the 2D error between the feature's center and the screen's center (Fig.~\ref{fig:centroid_tracking}). The center of the feature was tracked using a modification to the \quotes{stable grasp} object \cite{avinash_pickup_2019}. Four chamfered edges were added to the object, each serving as a mount for a unique 7.5mm$\times$7.5mm ArUco marker as shown in Fig.~\ref{fig:centroid_tracking}a.

ArUco markers \cite{garrido2014automatic} were tracked using OpenCV 4.8, and the marker with the largest perimeter was taken as the reference. Knowing the 2D corners of the ArUco marker in the auxiliary camera frame and the corresponding 3D relationship of the corners, the pose of the ArUCo marker with respect to the camera, ${}^{cam}\boldsymbol{T}_{ar}$, was computed by solving the Perspective-n-Point problem using Levenberg-Marquardt optimization \cite{marquardt1963algorithm}. The homogeneous transformation from the ArUco frame to the centroid of the tracked feature, ${}^{ar}\boldsymbol{T}_{cent}$, was found by matching the unique identification number of the ArUco marker to the predefined geometry of the \quotes{stable grasp} object. The real-time pose of the feature's centroid with respect to the auxiliary camera was then calculated as %${}^{cam}\boldsymbol{T}_{cent}=\{{}^{cam}\boldsymbol{R}_{cent},{}^{cam}\boldsymbol{t}_{cent}\}={}^{cam}\boldsymbol{T}_{ar}{}^{ar}\boldsymbol{T}_{cent}$
${}^{cam}\boldsymbol{T}_{cent}={}^{cam}\boldsymbol{T}_{ar}{}^{ar}\boldsymbol{T}_{cent}$ as shown in Fig.~\ref{fig:centroid_tracking}b. Finally, the centroid's 3D position %, ${}^{c}\boldsymbol{t}_{cent}$,
was projected onto the 2D auxiliary camera coordinate system using the methods described in \cite{corke2023robotics_cv} and the camera's extrinsic parameters described in Section \ref{subsec:video_processing}. The centroid of the tracked feature --- the ring of the surgical training task --- was projected to both the left and right camera planes of the autonomous stereo vision system.

Two participants navigated the surgical training task to evaluate the ability of the controller to keep the feature's centroid in the center of the camera frame. The experiment consisted of 591 seconds of data on the position of the feature's centroid $(u,v)$ in the left and right frames.

To evaluate the tracking accuracy, the mean Euclidean error (L2 Error) between the centroid's $(u,v)$ position and the center of the camera frame was calculated in pixels. The L2 Error was normalized by the screen's diagonal length, yielding a percentage error. Additionally, the mean component-wise error of the centroid along the $u$-coordinate ($u$ Error) and $v$-coordinate ($v$ Error) of the camera was calculated as the absolute difference between the projected centroid position and the screen center. $u$ Error and $v$ Error were also normalized as percentages of the screen width and height, respectively.

Fig.~\ref{fig:centroid_tracking}c. and Fig.~\ref{fig:centroid_tracking}$d$. show density scatter plots of the $(u,v)$ pixel locations of the centroid with the \quotes{$\boldsymbol{\times}$} indicating the center of the camera's coordinate system. Fig.~\ref{fig:centroid_tracking}d. through Fig.~\ref{fig:centroid_tracking}h. indicate $u$ Error and $v$ Error as a function of the experiment time. Red arrows denoted by (i) in Fig.~\ref{fig:centroid_tracking}d. and Fig.~\ref{fig:centroid_tracking}g. indicate peaks in the $u$ Error across both cameras at corresponding time points. Likewise, the purple arrows (ii) in Fig.~\ref{fig:centroid_tracking}e. and Fig.~\ref{fig:centroid_tracking}h. indicate corresponding $v$ Errors. 

Table \ref{centroidtracking_table} summarizes the L2 and component-wise ($u,v$) errors of the autonomous camera system in tracking the feature's centroid. Average L2 Error across the left and right cameras was 15.06 $\pm$ 7.95\%. The average $u$ Error for the cameras was 14.26 $\pm$ 10.48\%, and the average $v$ Error was 12.13 $\pm$ 10.12\%. 
%The feature is visible in the left camera 95.52\% of the time, and in the right camera 99.58\% of the time.
Additionally, the feature was visible by at least one camera 99.84\% of the time and by both 95.21\%.
\begin{table}[h!]
\begin{center}
\caption{Centroid Tracking Performance}
\label{centroidtracking_table}
\begin{tabular}{lll}
\hline\hline
      & \textbf{Left Camera} & \textbf{Right Camera} \\
\hline
L2 Error (pix) & 67.05 $\pm$ 48.12 & 125.29 $\pm$ 34.32 \\
L2 Error (\%) & 10.48 $\pm$ 7.52 & 19.58 $\pm$ 5.36\\
$u$ Error (pix)& 36.59 $\pm$ 36.26 & 108.81 $\pm$ 42.75\\
$u$ Error (\%)& 7.15 $\pm$ 7.08 & 21.25 $\pm$ 8.35\\
$v$ Error (pix)& 49.00 $\pm$ 41.91 & 44.18 $\pm$ 35.45\\
$v$ Error (\%)& 12.76 $\pm$ 10.92 & 11.51 $\pm$ 9.23\\
Visibility (\%)& 95.52 & 99.58\\
\hline
\end{tabular}
\end{center}
\end{table}

\subsection{Constrained Inverse Kinematics Performance}
\label{subsec:IK_performance}

 Constrained inverse kinematics performance was quantified over two trials from two participants completing the single-handed wire-chaser task, consisting of 14,720 optimizations over 2371 seconds. Table~\ref{table:optimization_performance} holds the results of this analysis, with the same error metrics used as in Section \ref{subsec:overall_system_performance}. 
 
\begin{table}[h!] 
\begin{center}
\caption{Constrained Inverse Kinematics Performance}
\label{table:optimization_performance}
\begin{tabular}{@{\hspace{4pt}}l@{\hspace{4pt}}l@{\hspace{8pt}}|@{\hspace{8pt}}l@{\hspace{4pt}}l@{\hspace{4pt}}}
\hline\hline
$VVA$ ($^{\circ}$) &  4.38 $\pm$ 2.11 & $AN$ ($^{\circ}$) &  10.10 $\pm$ 24.22 \\
$FD$ (mm) & 1.95 $\pm$ 5.66 & $PN$ (mm)& 13.91 $\pm$ 17.97\\
$PF$ ($^{\circ}$) & 0.03 $\pm$ 0.05 & $LT$ (s) & 0.019 $\pm$ 0.01 \\
\hline
\end{tabular}
\end{center}
\end{table}

\subsection{Usability Analysis}
\label{subsec:usability_analysis}
\begin{figure}[h!]
\begin{center}    \includegraphics[width=0.9\columnwidth,trim={0.55cm 7.85cm 25.25cm 1.5cm},clip]{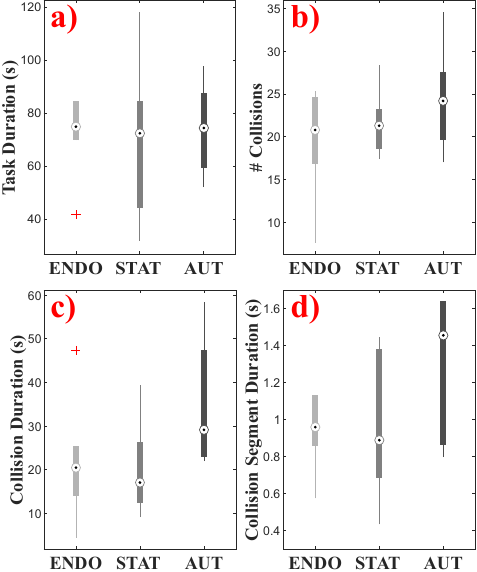}
    \caption{Quantitative results of usability analysis ($N=6$) comparing: 1) teleoperating from endoscopic camera only (ENDO), 2) teleoperating from stationary auxiliary camera (STAT), 3) teleoperating from AutoCam (AUT). \textbf{a)} Time taken to complete wire-chaser task; \textbf{b)} Number of collisions; \textbf{c)} Total time wire touched ring; \textbf{d)} Average duration of each collision.}
    \label{fig:PilotResults_Quantitative}
\end{center}
\end{figure}

Six participants completed the single-handed wire-chaser task under the ENDO, STAT, and AUT conditions described in Section \ref{subsec:experimental_design}. The quantitative results of this pilot study are illustrated in Fig. \ref{fig:PilotResults_Quantitative}, and there was no statistical significant difference between groups across all performance metrics. The autonomous camera (AUT) group completed the task in 74.26 $\pm$ 17.35 seconds, which was faster than the endoscopic (ENDO) group (79.82 $\pm$ 30.09 s) and slower than the stationary camera (STAT) group (70.55 $\pm$ 30.63 s). The number of collisions were similar with AUT having 24.53 $\pm$ 6.44 collisions compared to the 19.33 $\pm$ 6.60 and 21.70 $\pm$ 3.89 for ENDO and STAT, respectively. The collision duration --- the total time the wire touched the ring --- was greater for the experimental group (34.85 $\pm$ 14.76 seconds) compared to the ENDO (22.08 $\pm$ 14.47 s) and STAT (20.28 $\pm$ 11.05 s) groups. Additionally, no statistical difference was found for the duration of collision segments --- the average time users dragged the ring along the wire --- with the AUT group having 1.56 $\pm$ 0.84 s, the ENDO group having 1.07 $\pm$ 0.46 s, and the STAT group having 0.95 $\pm$ 0.40 s segment durations.

NASA-TLX results are given in Appendix~\ref{appendix:usability_analysis}, with no significant differences between the three conditions.

\section{Discussion}
We introduce a markerless approach to autonomously control a pick-up camera with the da Vinci surgical robot, and rigorously quantify the system's performance in a dry-lab setting. While earlier studies have used reinforcement learning to control an auxiliary viewpoint \cite{ji2018learning,subedi_smoothness_2023,su_deep_2022,su_multicamera_2020,su_multicamera_2021}, these methods do not enable 6 DOF camera motion, have been only tested in simulation, and do not account for workspace or joint limit constraints. Another approach has used U-Net-based deep learning with kinematic constraints to adjust the field of view of a microscopic camera \cite{lin_autonomous_2024}. This method, however, does not account for camera orientation and only evaluates performance in 2D trajectories. Our paper presents a hierarchical, 6 DOF, autonomous controller that considers both workspace and robot constraints and is implemented for real-time use on a da Vinci system with the dVRK.

When evaluating the overall system on a dry-lab surgical training task, we found that our autonomous controller could track the feature accurately while accounting for workspace and robot constraints. The camera pointed toward the center of the feature with a small viewing vector angle error of 1.71 $\pm$ 1.54\textdegree. Seeing that the Raspberry Pi V2 Camera has a horizontal FoV of 62.2\textdegree \;and a vertical FoV of 48.8\textdegree, this viewing angle error is only 2.7-3.5\% of the FoV. The controller kept the feature within the desired distance by 4.98 $\pm$ 5.00 mm. The ability of the controller to maintain a dynamic feature at a constant distance is important, especially when AutoCam is extended to tracking motion of tissue surfaces for scanning and registration applications \cite{schmidt2024surgical}. We also found that the controller kept the camera parallel to the floor with a small error of 0.34 $\pm$ 0.21\textdegree. This demonstrates AutoCam’s ability to orient the camera relative to key surgical planes, such as a horizontal view of the liver during abdominal procedures  \cite{zheng2017your}. 

The ability of users to view the salient feature was also assessed by finding the on-screen distance of the feature's centroid from the center of the 2D image plane (Fig.~\ref{fig:centroid_tracking}). When combining the results of the left and right cameras in Table~\ref{centroidtracking_table}, the controller kept the feature at the center of the image plane with an error of 15.03 $\pm$ 7.96\% of the screen size.  When looking at the red arrows in Fig.~\ref{fig:centroid_tracking}d. and Fig.~\ref{fig:centroid_tracking}g., we see that areas of higher error in the left camera's $u$-axis corresponds to areas of lower error for the right camera. This is because it is impossible to keep the centroid at the center of both cameras in a stereo pair when the cameras optical axes are collinear. In addition, both the $u$-axis and $v$-axis errors have a periodic waveform. This is caused by the controller filtering its pose in response to rapid motions of the feature caused by workspace constraints or abrupt turns in the feature's path. Although the tracking error is variable, it rarely exceeds 40\% of the screen size, indicating that the feature is kept near the center of the screen as seen in Fig.~\ref{fig:centroid_tracking}c. and d. 

Importantly, the feature was kept within the view of at least one camera 99.84\% of the time, evidencing the controller's ability to maintain reliable on-screen visualization. The RL approach proposed in \cite{su_multicamera_2020,su_multicamera_2021,subedi_smoothness_2023} attains a visibility of 71.0\% in simulation and the deep learning approach in \cite{lin_autonomous_2024} achieves 94.1\% visibility. Compared to these state-of-the art methods, our approach achieves higher visibility while also controlling the full 6 DOF of the camera in real-time and accounting for system and workspace constraints. Ultimately, our system ensures that salient objects are maintained within the camera viewpoint, thus minimizing poor visualization which can compromise patient safety \cite{millan_scoping_2021}.

The commanded pose of the auxiliary camera was compared to the naive pose --- the pose computed if workspace and joint limit constraints were not considered. Fig.~\ref{fig:actual_to_naive}a. and b. show peaks in the orientation and position error when constraint conditions were triggered. Accounting for factors such as the No-Go Zone and Joint Limits led to deviation away from a completely orthogonal view to the salient feature. However, ensuring the camera does not violate these constraints is important especially in surgeries such as lung tumor resections where strict \quotes{no-fly zones} are defined \cite{rocco2023commentary}. The controllers ability to avoid crossing the no-go zone is exemplified in Fig.~\ref{fig:actual_to_naive}c., where the actual trajectory does not cross the horizontal plane representing an organ surface such as the pancreas. When constraint conditions are removed from the experiment, the controller tracks the geometrically computed, naive, pose with an orientation error of 6.58 $\pm$ 6.72\textdegree \;and position error of 4.63 $\pm$ 3.03 mm. This demonstrates the ability of the controller to maintain an ideal view of the feature while simultaneously accounting for obstacles in the surgical scene.

To account for boundary conditions, we present a novel hierarchical controller that incorporates constrained inverse kinematics with loss function given in \eqbracketref{optim_eq1}. In addition to evaluating overall system performance, the performance of this constrained optimization was assessed independently (Table \ref{table:optimization_performance}). The controller was both computationally efficient (19 $\pm$ 10 ms average solution time) and accurate, maintaining a viewing vector angle error of 4.38 $\pm$ 2.11\textdegree \;and feature distance error of 1.95 $\pm$ 5.66 mm. Our optimization performs much faster than the state-of-the-art RL approach in \cite{su_multicamera_2020}, which has a loop speed of 113ms. Furthermore, their approach does not account for mechanical constraints and was determined to have a visibility score of 66.1\% in \cite{subedi_smoothness_2023}, which is much lower than our 99.58\% visibility. This suggests that our optimization method outperforms \cite{su_multicamera_2020} for both speed and accuracy while robustly handling joint limit constraints. 

A separate pilot study (Fig. \ref{fig:PilotResults_Quantitative}) showed that novices could teleoperate from the AutoCam viewpoint with similar proficiency to using the ECM or a stationary auxiliary camera. This suggests that an autonomous secondary view can be integrated into RAMIS training without degrading novice performance. Rather than aiming to improve novice performance, our pilot study evaluates whether AutoCam can match existing single-camera teleoperation methods. With more specific training, such as that received by expert surgeons, we anticipate a framework like AutoCam could integrate into surgical workflows. Furthermore, it has been shown that when the ECM is obscured due to instrument occlusions or lens fouling, operative time increases by 19.3\%, and there are higher rates of surgical errors \cite{nabeel2024assessing}. AutoCam provides a method to overcome such occlusions by improving spatial awareness and allowing surgeons to monitor the surgical scene while cleaning the ECM lens. This aligns with the long-term vision of the RAMIS community to develop semi-autonomous surgical robots and path planning algorithms for better camera control \cite{millan_scoping_2021,d2021accelerating}.

Although our system was shown to work equivalently to other teleoperation methods, tracked orientation robustly, and had a high visibility score of 99.84\%, there are limitations in our methods and their evaluation that can be addressed in future studies. Firstly, ArUco markers are segmented with 98\% pixel accuracy \cite{xavier2017accuracy}, meaning there are small inaccuracies in the baseline measurement that may impact the centroid tracking results. Future studies can cross-validate with other tracking methods, such as using Northern Digital\textregistered \;infrared tracking technology. Another area for improvement is to introduce a mechanism to seamlessly transition between teleoperating from the external camera frame to the endoscope frame. Using eye-gaze or another input signal to switch between teleoperation frames could overcome the negative impact of control-viewpoint misalignment \cite{chintamani2009improved}. Additionally, while our controller accurately tracks the feature compared to a vision baseline, future research could implement different control frameworks on the dVRK including reinforcement learning, model predictive control, and end-to-end optimization. These methods could be compared to our hierarchical approach for performance differences including computational efficiency, viewing accuracy, and generalizability to other surgical scenes. 

Further studies can also combine our autonomous camera control framework with advanced visual tracking and perception algorithms. Using methods such as the recurrent implicit neural graph in \cite{schmidt2022recurrent} to enable AutoCam to track deformable tissue aligns with broader goals to automate camera motion in surgery \cite{salcudean2022robot}. Machine learning models could also be trained on datasets such as SurgPose \cite{wu2025surgpose} for markerless tool tracking. Additionally, depth mapping via stereo image matching could allow dynamic updates to the no-go zone, promoting greater autonomy and contextual awareness. Finally, although AutoCam is shown in a six-participant user study to integrate into dry-lab surgical training, larger studies are needed to explore the specific benefits of an autonomous auxiliary viewpoint on RAMIS training and intraoperative performance.

\section{Conclusion}
Recent studies suggest that multiple viewpoints provide additional information on scale, depth, and spatial relations during robot teleoperation \cite{praveena2022understanding}, which enhances surgical performance and global awareness during RAMIS \cite{pandya2014review}. Although hardware solutions such as the \quotes{pick-up} camera \cite{avinash_pickup_2019} provide an additional viewpoint in RAMIS, most systems require manual control, adding more workload to the surgical team. To address this undesirable additional burden, our paper introduces a novel method for an autonomously-guided external camera. By enabling 6 DOF camera control, accounting for workspace and robot constraints, and validating our method with the da Vinci surgical robot, our system overcomes implementation challenges of previous autonomous approaches. The code for our approach is available at: \href{https://github.com/AlexandreBanks6/RCL_Autocam_Public.git}{RCL\_Autocam\_Public.git}. We quantified technical feasibility by showing that our approach maintained the salient feature's centroid near the screen center with a 15.03 $\pm$ 7.96\% error, despite obstacles and joint limits.  Additionally, the system kept the feature within view 99.84\% of the time, ensuring that salient points will be tracked reliably. Furthermore, deviation from the ideal orthogonal view was only 6.58 $\pm$ 6.82\textdegree, giving the user a continuously optimized view of the scene. Future studies can further explore AutoCam's impact on novice performance, optimize it for intraoperative use, and combine the autonomous controller with tissue tracking. Our markerless, constraint-aware, control approach integrates with the da Vinci surgical robot and paves the way for hands-free auxiliary viewpoint control in RAMIS.

\section*{Acknowledgments}
The authors gratefully acknowledge support from the NSERC Canada Graduate Scholarships, and both the NSERC Discovery Grant with the Canadian Department of National Defense supplement and the C.A. Laszlo Biomedical Engineering Chair held by Professor Salcudean. We  also thank Intuitive Surgical and Johns Hopkins University for support with the da Vinci Research Kit. The authors report no project-related conflicts of interest.
\bibliographystyle{IEEEtran}
\bibliography{Autocam_Bibliography}
\newpage
%\vspace*{-0.4in}
\begin{IEEEbiography}[{\includegraphics[width=1in,height=1.25in,clip,keepaspectratio]{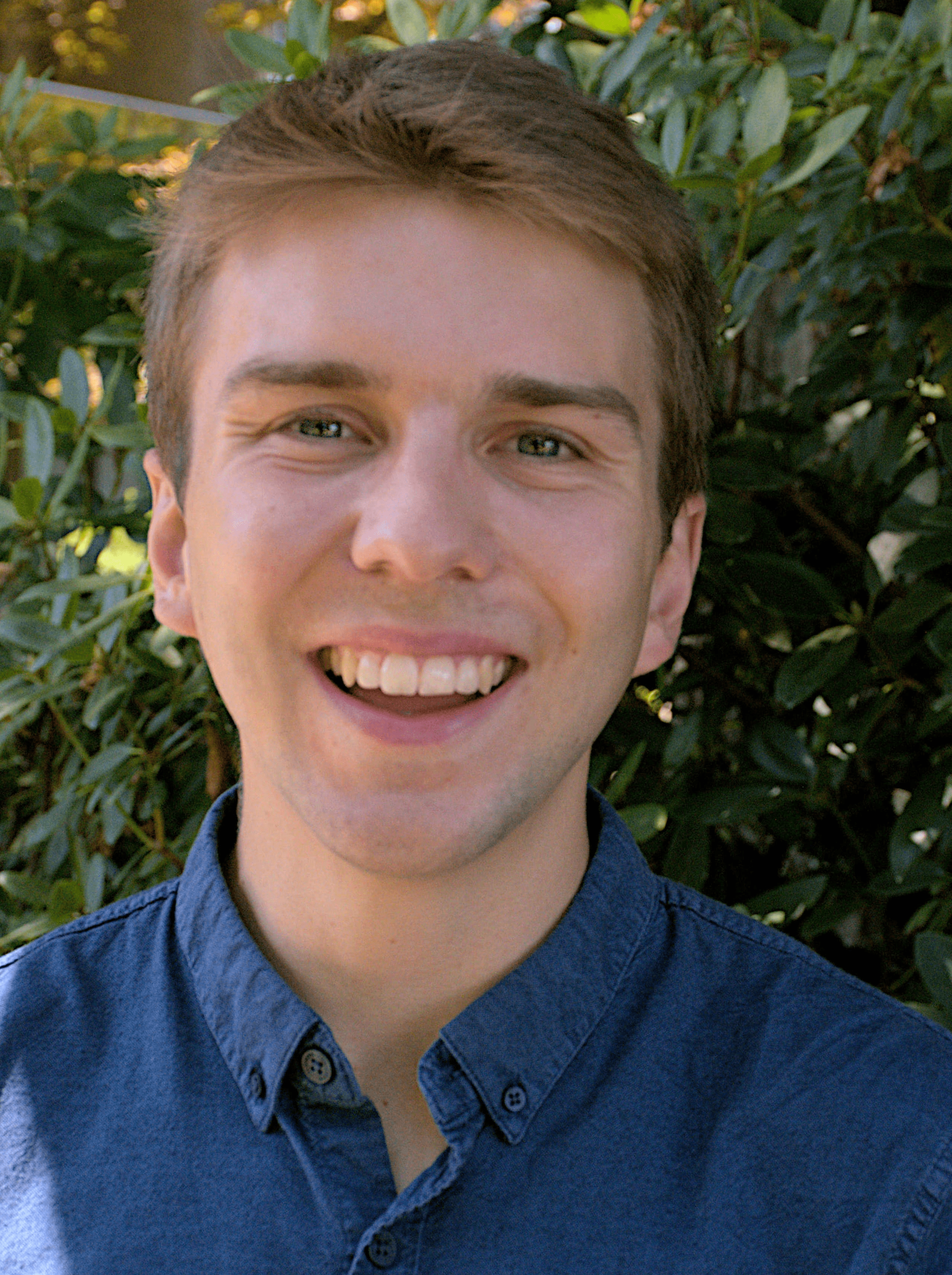}}]{Alexandre Banks} was born in Saint John, NB, Canada. After graduating with a B.Sc. in Electrical and Computer Engineering from the University of New Brunswick (UNB), he moved to the University of British Columbia (UBC) where he is pursuing an M.A.Sc. in Biomedical Engineering. Alexandre completed research internships in rehabilitation robotics at the UNB Institute of Biomedical Engineering in 2020 and 2021, and from 2022-2025 worked in surgical robotics at the UBC Robotics and Control Laboratory. In 2025, Alexandre completed an internship at Oxford University on computer vision for fetal ultrasound.
\end{IEEEbiography}
\vspace*{-0.3in}
\begin{IEEEbiography}
    [{\includegraphics[width=1in,height=1.25in,clip,keepaspectratio]{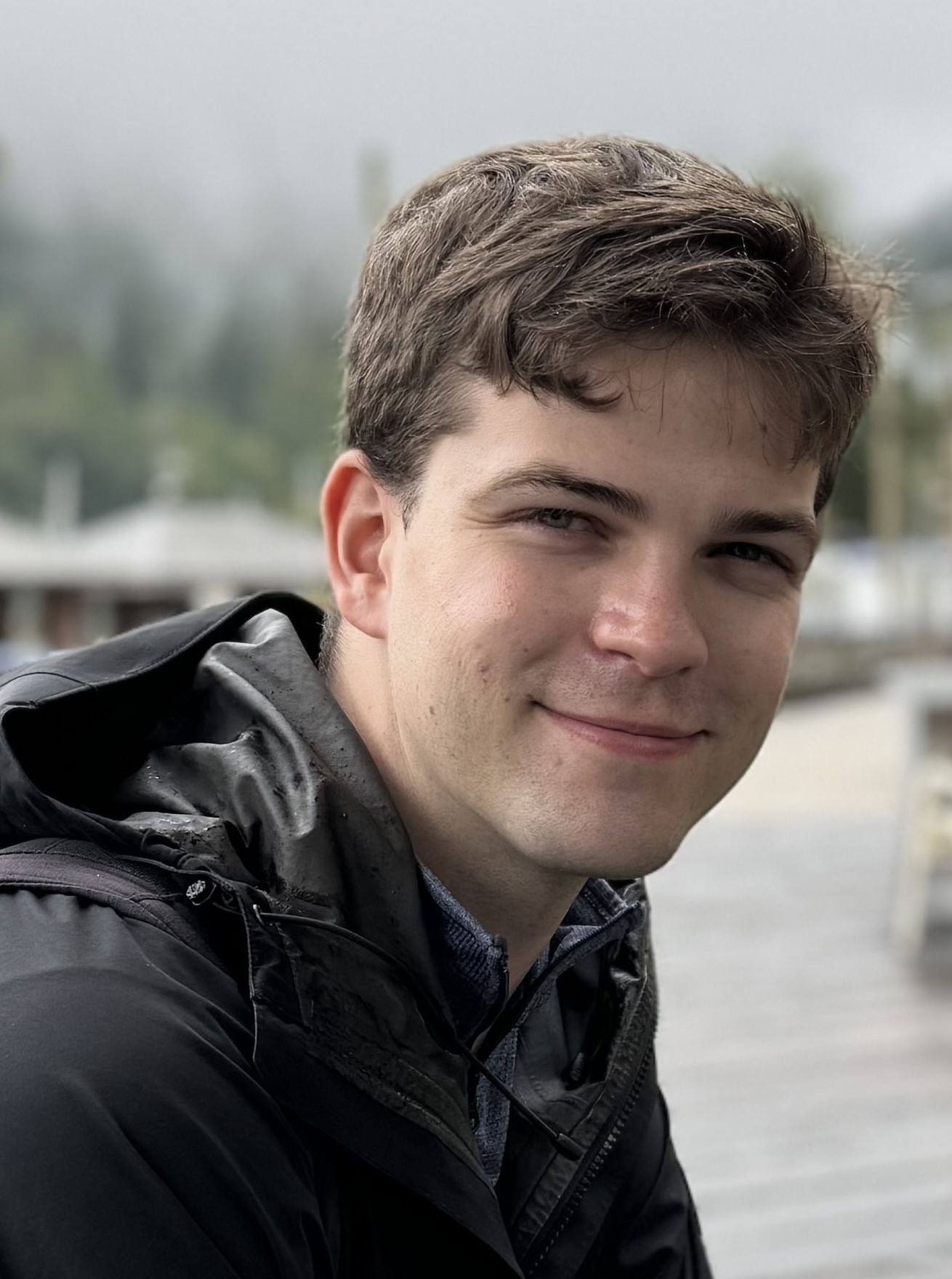}}]{Randy Moore} was born in Calgary, AB, Canada, in 1999. Following the completion of a B.Sc. in Electrical and Computer Engineering at the University of Calgary (UofC), he completed a M.A.Sc in Electrical and Computer Engineering on image guidance in surgical robotics at the Robotics and Control Laboratory at the University of British Columbia. Randy previously worked with Project NeuroArm of the UofC on neurosurgical robotics and now works on image-guided applications in the aerospace industry.  
\end{IEEEbiography}
\vspace*{-0.3in}
\begin{IEEEbiography}
    [{\includegraphics[width=1in,height=1.25in,clip,keepaspectratio]{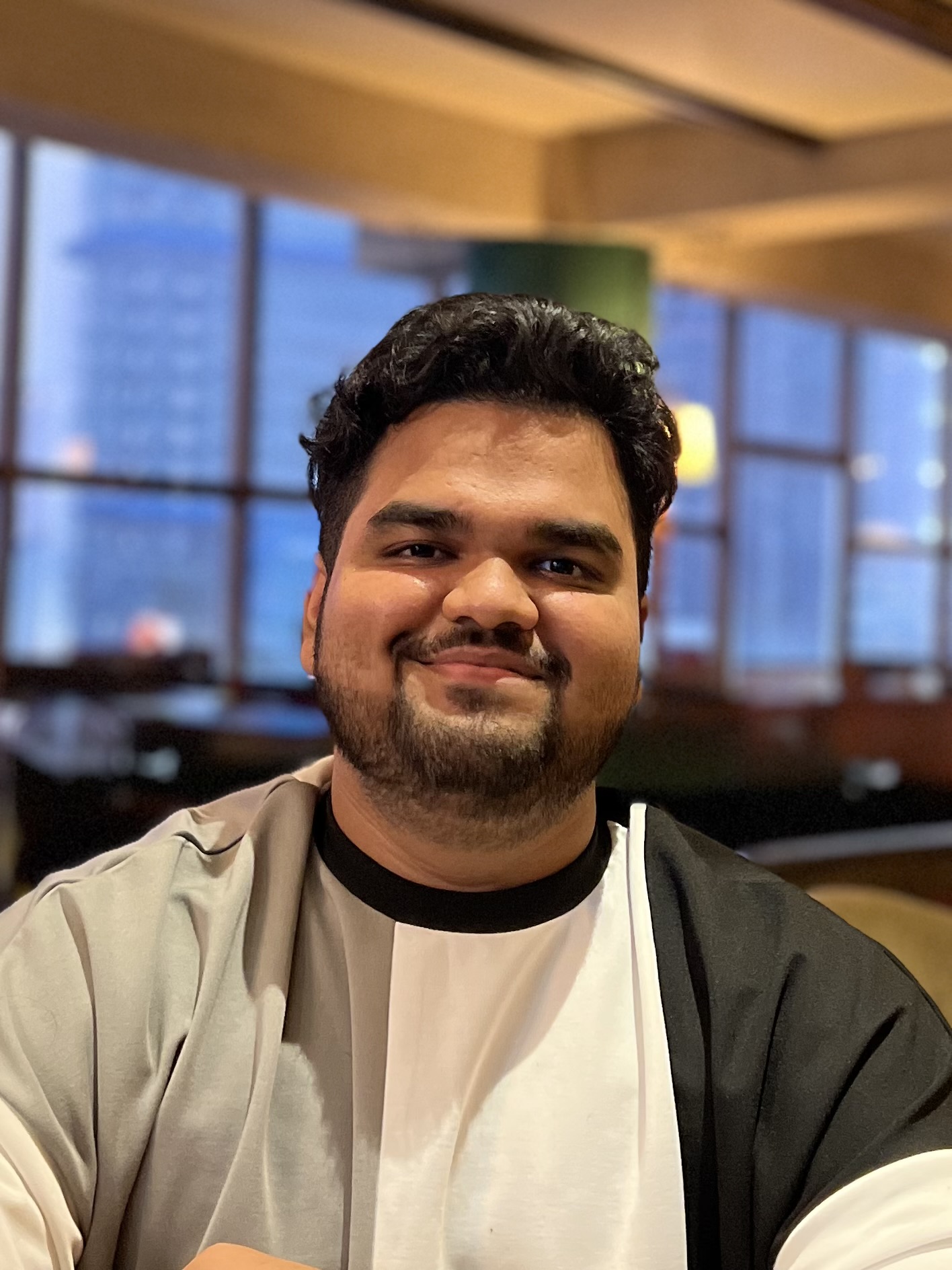}}]{Sayem N. Zaman} was born in Dhaka, Bangladesh. He received his B.A.Sc. degree in Engineering Physics from the University of British Columbia (UBC), Vancouver, BC, Canada, in May 2024. During his undergraduate studies, he completed a research internship in 2020 at the Center for Aerospace Research, University of Victoria. From 2021-2024, he was involved in several research projects in surgical robotics at the Robotics and Controls Laboratory, UBC.  
\end{IEEEbiography}
\vspace*{-0.3in}
\begin{IEEEbiography}
    [{\includegraphics[width=1in,height=1.25in,clip,keepaspectratio]{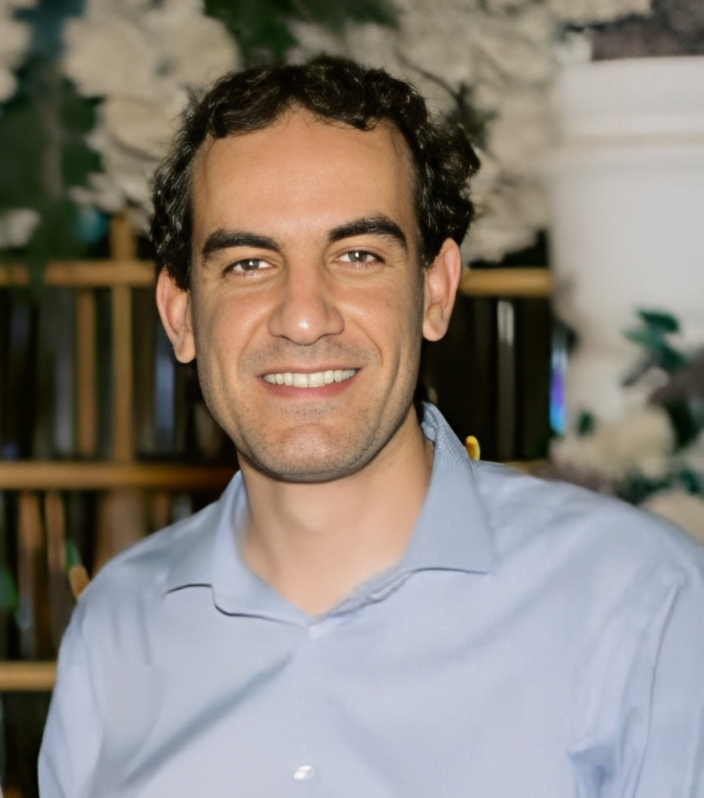}}]{Alaa Eldin Abdelaal} is a postdoctoral scholar at Stanford University. He received his PhD in Electrical and Computer Engineering from the University of British Columbia (UBC) in December 2022. He was also a visiting graduate scholar at the Computational Interaction and Robotics Lab at Johns Hopkins University. He holds a M.Sc. in Computing Science from Simon Fraser University and a B.Sc. in Computer and Systems Engineering from Mansoura University in Egypt. His research interests are at the intersection of automation and human-robot interaction for human skill augmentation and decision support with application to surgical robotics. His research has been funded by a Vanier Canada Graduate Scholarship, an NSERC Postdoctoral Fellowship, Intuitive Surgical Inc., and the Institute for Human-Centered Artificial Intelligence at Stanford University.
\end{IEEEbiography}
\vspace*{-0.3in}
\begin{IEEEbiography}
[{\includegraphics[width=1in,height=1.25in,clip,keepaspectratio]{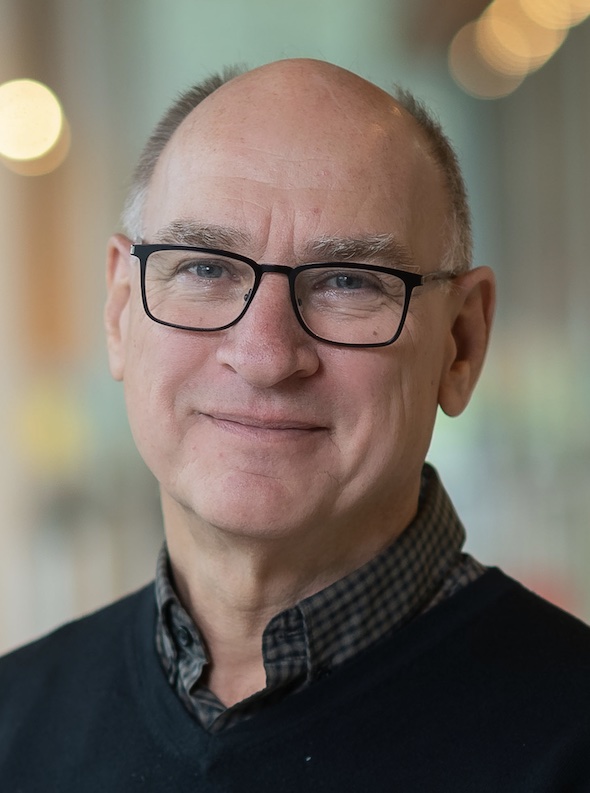}}]{Septimiu E. Salcudean} (Fellow, IEEE) was born in Cluj, Romania. He received the B.Eng. (Hons.) and M.Eng. degrees in electrical engineering from McGill University, Montreal, QC, Canada, in 1979 and 1981, respectively, and the Ph.D. degree in electrical engineering from the University of California at Berkeley, Berkeley, USA in 1986. He was a Research Staff Member with the IBM T. J. Watson Research Center from 1986 to 1989. He then joined UBC where he is currently a Professor with the Department of Electrical and Computer Engineering and the School of Biomedical Engineering. He holds the C.A. Laszlo Chair of Biomedical Engineering. He has been a Co-Organizer of the Haptics Symposium, a Technical Editor and Senior Editor of the IEEE Transactions on Robotics and Automation, and on the program committee of the ICRA, MICCAI, and IPCAI Conferences. He is currently on the Steering Committee of the IPCAI Conference and on the editorial board of The International Journal of Robotics Research. He is a Fellow of MICCAI and of the Canadian Academy of Engineering.
\end{IEEEbiography}
\newpage
%Appendices
\appendices
\normalfont
\section{Equation Definitions}
\label{appendix:eqdefs}
\subsection{Huber Loss}
The definition of the huber loss, $h\{\}$ used in our paper is given as
\begin{equation*}
 {h\{x,\delta\}} =
 \begin{cases}
 \frac{1}{2}x^2,&{\text{if}}\ |x|\leq\delta,\\
 {\delta(|x|-\frac{1}{2}\delta),}&{\text{otherwise.}}
 \end{cases}
\end{equation*}

Delta values used in the constrained IK objective function \eqbracketref{optim_eq1} are given in Appendix \ref{appendix:parametervals} as, $\delta_1$, $\delta_2$ and $\delta_3$.

\section{Parameter Values}
\label{appendix:parametervals}
The parameters and hyperparameters were determined empirically by systematically incrementing one parameter while holding all others constant. Computational speed and the error metrics in Table~\ref{table:overall_system_errors} were considered during this process. Parameter selection occurred prior to the experiments in the accompanying paper and all parameters were held constant throughout the user studies.

\begin{table}[h!]
\begin{center}
\caption{Parameter Values Implemented on the dVRK}
\label{table:appendix_parameter values}
\resizebox{\columnwidth}{!}{
\begin{tabular}{lll}
\hline\hline
Parameter Description      & Parameter Symbol & Parameter Value \\
\hline
Regularization Terms & $w_1$, $w_2$, $w_3$, $w_4$, $w_5$ & 15.0, 30.0, 25.0, 2.0, 0.5 \\
Huber Loss Thresholds      & $\delta_1$, $\delta_2$, $\delta_3$ & 0.01, 0.17, 0.02 \\
\hline
\end{tabular}
}
\end{center}
\end{table}

\section{Usability Analysis NASA-TLX Results}
\label{appendix:usability_analysis}

\begin{figure}[h!]
\begin{center}    \includegraphics[width=\columnwidth,trim={0.5cm 2.4cm 16cm 1cm},clip]{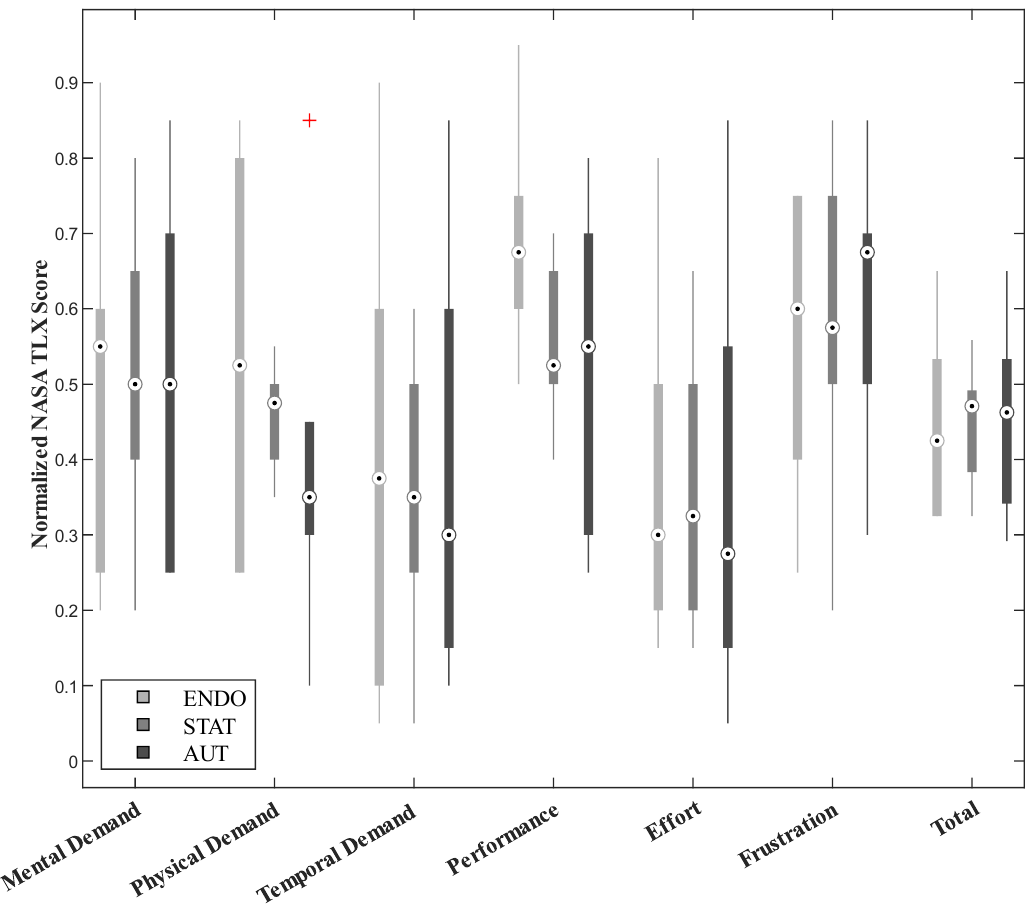}
    \caption{Qualitative NASA-TLX results of usability analysis (N=6) comparing: 1) teleoperating from endoscopic camera only (ENDO), 2) teleoperating from stationary auxiliary camera (STAT), 3) teleoperating from AutoCam (AUT).}
    \label{fig:PilotResults_Qualitative}
\end{center}
\end{figure}

Qualitative results of the pilot study (N=6) where participants completed the single-handed wire-chaser task are given in Fig. \ref{fig:PilotResults_Qualitative}. Participants completed five trials of each condition (ENDO, STAT, and AUT), and their self-reported feedback was assessed after every condition using the NASA-TLX questionnaire. No statistically significant differences are found between conditions. Notably, physical and temporal demand was lower for the autonomous auxiliary camera condition compared to using only the endoscope (ENDO) or the endoscope plus a stationary auxiliary camera (STAT). Effort and mental demand were similar across all conditions. The autonomous camera condition also led to lower self-reported performance than the ENDO condition, and higher frustration than both the ENDO and STAT conditions.

\section{Confidence Intervals}
\label{appendix:confidence_intervals}
The confidence intervals for each variable of interest have been plotted for each of the three groups. The means for each variable fall within the 95\% confidence interval for all other conditions, suggesting that there could be no statistically significant difference between them. 
\begin{figure}[h!]
\begin{center}    \includegraphics[width=\columnwidth,clip]{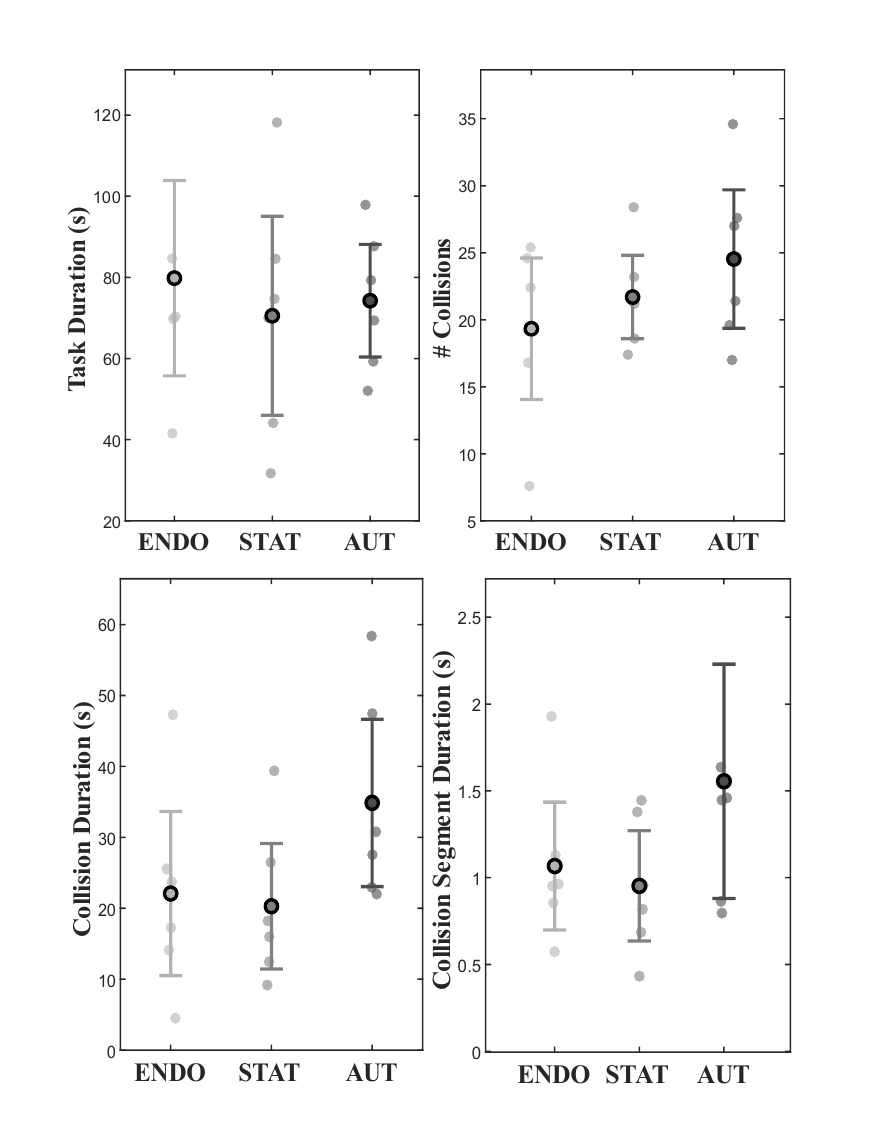}
    \caption{Confidence Intervals (95\%) for User Study Groups}
    \label{fig:PilotResults_ConfidenceIntervals}
\end{center}
\end{figure}

\vfill

\end{document}